\documentclass{applemlr}

\usepackage{amsmath}
\usepackage{enumerate}
\usepackage{algorithm}
\usepackage{algpseudocode}
\usepackage{amsfonts}
\usepackage{amsthm}
\usepackage{cleveref}
\usepackage{diagbox}
\usepackage{colortbl}
\usepackage{amssymb}
\usepackage{xspace}
\usepackage{wrapfig}
\usepackage{adjustbox}
\usepackage{tabularx}
\usepackage{booktabs}
\usepackage{mathtools}
\usepackage{tikz}
\usepackage{enumitem}
\usepackage{silence}
\usepackage{dsfont}
\usepackage[table]{xcolor}
\usepackage[dvipsnames]{xcolor}
\usepackage{multirow}
\usepackage{makecell}
\usepackage{xfakebold}

\usepackage{amsmath,amsfonts,bm}

\def\eqref#1{equation~\ref{#1}}

\def\1{\bm{1}}

\DeclareMathAlphabet{\mathsfit}{\encodingdefault}{\sfdefault}{m}{sl}
\SetMathAlphabet{\mathsfit}{bold}{\encodingdefault}{\sfdefault}{bx}{n}

\definecolor{textgray}{HTML}{6E6E73}
\usetikzlibrary{positioning, calc}
\usetikzlibrary{decorations.pathmorphing}

\makeatletter
\patchcmd{\wrong@fontshape}{\@gobbletwo}{}{}{}
\makeatother
\numberwithin{equation}{section}
\makeatletter
\AtBeginDocument{
  \urlstyle{sf}
  
}
\makeatother

\definecolor{light}{RGB}{125, 125, 125}
\crefname{tcb@cnt@pbox}{code}{code}
\Crefname{tcb@cnt@pbox}{Code}{Code}
\crefname{assumption}{assumption}{assumption}
\Crefname{assumption}{Assumption}{Assumptions}

\newtcolorbox[auto counter]{pbox}[2][]{
  colback=white,
  title=Code~\thetcbcounter: #2,
  #1,fonttitle=\sffamily,
  fontupper=\sffamily,
  arc=2pt,
  colframe=bgcolor,
  coltitle=fgcolor,
  colbacktitle=bgcolor,
  toptitle=0.25cm,
  bottomtitle=0.125cm
}

\makeatletter
\newcommand\applefootnote[1]{%
  \begingroup
  \renewcommand\thefootnote{}%
  \renewcommand\@makefntext[1]{\noindent##1}%
  \footnote{#1}%
  \addtocounter{footnote}{-1}%
  \endgroup
}
\makeatother

\definecolor{cverbbg}{gray}{0.90}

\usepackage{mathrsfs}
\usepackage[textsize=tiny]{todonotes}

\newcommand{\bogdan}[1]{}
\newcommand{\asn}[1]{}
\newcommand{\as}[1]{}
\newcommand{\mkn}[1]{}

\usepackage{xcolor}      %

\usepackage{url}

\usepackage{bbm}
\usepackage{xspace}
\usepackage{enumitem}
\usepackage{wrapfig}
\usepackage{float}

\newcommand{\Method}{Strategy-Guided Exploration\xspace}
\newcommand{\method}{SGE\xspace}
\newcommand{\passk}{pass@$k$\xspace}

\usepackage{listings}

\definecolor{applegray}{HTML}{F5F5F7}

\lstdefinestyle{rawtext}{
  basicstyle=\ttfamily\footnotesize,
  columns=fullflexible,
  keepspaces=true,
  breaklines=true,
  showstringspaces=false
}

\newtcolorbox{rawpanel}[1][]{%
  enhanced,
  arc=8pt,
  outer arc=8pt,
  boxrule=0.7pt,
  colframe=black,
  colback=applegray,
  top=10pt,
  bottom=10pt,
  left=12pt,
  right=12pt,
  #1
}

\title{Expanding LLM Agent Boundaries with Strategy-Guided Exploration}

\author{Andrew Szot}
\author{Michael Kirchhof}
\author{Omar Attia}
\author{Alexander Toshev}

\affiliation{Apple}

\abstract{
  Reinforcement learning (RL) has demonstrated notable success in post-training large language models (LLMs) as agents for tasks such as computer use, tool calling, and coding. However, exploration remains a central challenge in RL for LLM agents, especially as they operate in language-action spaces with complex observations and sparse outcome rewards. In this work, we address exploration for LLM agents by leveraging the ability of LLMs to plan and reason in language about the environment to shift exploration from low-level actions to higher-level language strategies. We thus propose Strategy-Guided Exploration (SGE), which first generates a concise natural-language strategy that describes what to do to make progress toward the goal, and then generates environment actions conditioned on that strategy. By exploring in the space of strategies rather than the space of actions, SGE induces structured and diverse exploration that targets different environment outcomes. To increase strategy diversity during RL, SGE introduces mixed-temperature sampling, which explores diverse strategies in parallel, along with a strategy reflection process that grounds strategy generation on the outcomes of previous strategies in the environment. Across UI interaction, tool-calling, coding, and embodied agent environments, SGE consistently outperforms exploration-focused RL baselines, improving both learning efficiency and final performance. We show that SGE enables the agent to learn to solve tasks too difficult for the base model.

}

\metadata[Correspondence]{\sffamily Andrew Szot: \url{a_szot@apple.com}; Alexander Toshev: \url{toshev@apple.com}}

\date{\sffamily\today}

\begin{document}

\maketitle

\section{Introduction}

Large language models (LLMs) are a promising foundation for building agents across a wide range of downstream tasks such as computer use \citep{cua2025}, coding \citep{claudecode}, tool usage \citep{o3o4cite}, and robotics \citep{kim2024openvla}.
A key driver of progress in LLM agents is reinforcement learning (RL), which trains LLMs to autonomously act to solve tasks in external environments.
RL has shown the ability to teach LLMs to reason, self-correct, and interact with complex environments using simple, easy-to-specify rewards without human feedback \citep{guo2025deepseek}.

Exploration is a central challenge in training LLM agents with RL.
This challenge is amplified by sparse outcome rewards, which reward the agent only when it achieves the desired goal.
In domains such as video game playing \citep{silver2017mastering} or navigation \citep{wijmans2019dd}, RL can train superhuman policies starting from a base policy that performs poorly.
In contrast, RL for LLM agents operates in complex language action spaces such as code outputs or tool calls, and is initialized from a pretrained model that induces a concentrated policy over likely outputs.
As a result, RL for LLM agents predominantly samples from and refines behaviors already supported by the base model, limiting the agent’s ability to discover new successful trajectories.
Consistent with this, empirical evidence from prior work in non-agentic reasoning domains demonstrates that RL post-training struggles to learn new tasks and instead refines outputs to high-reward trajectories the model is already capable of generating \citep{wu2025invisible,yue2025does}.

To improve exploration in RL training for LLM agents, we introduce \Method (\method), which expands agent capabilities by exploring new tasks not solvable by the base model.
\method leverages the ability of LLMs to plan and reason in natural language about the agent’s environment.
Instead of treating exploration as sampling environment actions, \method first generates a language ``strategy" and then produces the environment action conditioned on that strategy.
The strategy is a concise and specific natural language description of what to do to make progress toward the goal.
We show that with \method, exploring in the space of strategies is easier than exploring in the space of actions, and that diverse strategies lead to better exploration in the environment, which increases the likelihood of success.
Since strategies are high-level language descriptions of intended outcomes, the LLM can use its reasoning abilities to generate many distinct strategies. 
Conditioning on these strategies then guides the LLM to produce action sequences consistent with the strategy that explore different environment outcomes.

\begin{figure*}[t!]
  \centering
  \includegraphics[width=\textwidth]{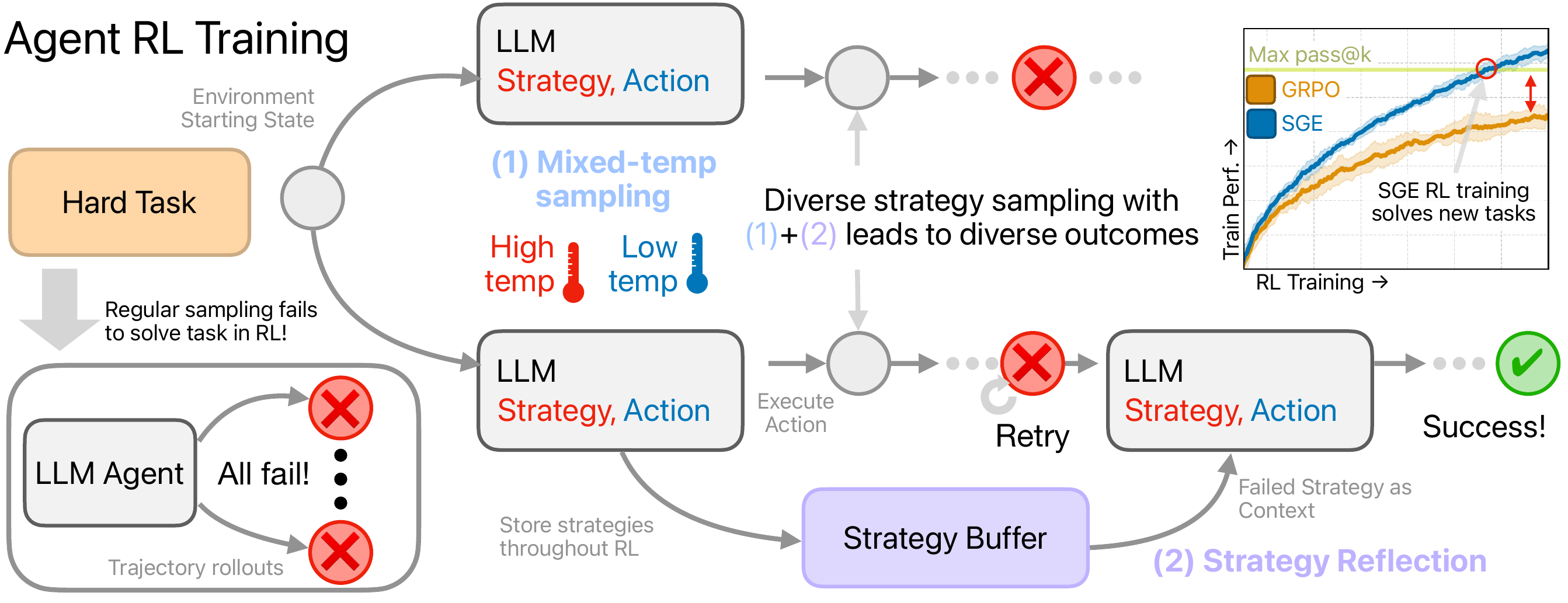}
  \caption{
    Overview of \Method (\method). 
    \method improves reinforcement learning (RL) training on hard agentic tasks for which the base model fails, even after many attempts (left of figure). 
    \method addresses this by having the LLM policy output a language ``strategy'' and then conditioning the action generation on this strategy.
    The reasoning capabilities of the LLM enable it to output diverse strategies through the techniques of: (1) mixed-temperature sampling, where strategy tokens are sampled with higher temperature than remaining tokens, and (2) strategy reflection, where strategies are generated to be distinct from other strategies executed earlier in RL training.
    This enables \method to explore to solve hard tasks that the base model is not capable of succeeding in (right of figure).
  }
  \label{fig:intro}
\end{figure*}

\method introduces several techniques to generate diverse strategies during RL training to improve exploration.
\method does more than simply prompt the LLM agent to first produce a strategy, and introduces techniques to maximize strategy diversity throughout training.
The first component of \method is mixed-temperature sampling, which samples the strategy at a higher temperature than the rest of the token sequence.
Higher temperature for action sampling often produces different actions that achieve the same environment outcome, such as clicking the same UI element at different positions.
In contrast, increasing temperature in the higher-level strategy language space generates strategies that correspond to different outcomes.
The second component of \method is strategy reflection, where the model reflects on successful and failed strategies from earlier in RL training.
This reflection helps ground strategy generation in the details of the environment dynamics and further diversify strategies from previous attempts.
See \Cref{fig:intro} for an overview of the \method method.

Our experiments show that \method solves challenging tasks during RL training that the base model cannot solve even after many attempts across the multi-step agentic environments of coding \citep{xia2025leetcodedataset}, UI control \citep{rawles2024androidworld}, tool-calling \citep{trivedi2024appworld}, and embodied agents \citep{szot2023large}.
\method also outperforms exploration-focused RL baselines, achieving higher final performance and better learning efficiency.
For example, in the multi-turn coding environment, the starting LLM solves 69\% of the problems \emph{at least once} out of 2048 attempts (pass@2048=0.69). 
The best performing RL baseline with a 64\% final success rate (pass@1=0.64) is unable to surpass this ceiling in task pass rate as it attains no learning signal for the remaining difficult problems.
\method, capable of exploring to solve harder problems, achieves a 73\% success rate.
These training gains then translate to better generalization on unseen tasks.
We demonstrate the broad applicability of \method by showing consistent improvements with \method across four diverse environments involving different observation and action spaces.
We also ablate the \method components of mixed-temperature sampling and strategy reflection, along with analyzing the impact of \method across LLM size.
We qualitatively analyze how the LLM is able to generate and use diverse strategies to solve new tasks that are unsolvable by the base model.
Overall, our experiments show how \method uses exploration to unlock scaling RL training to expand agentic capabilities on difficult tasks.

\section{Related Work} 

Exploration is a long-standing central challenge of RL.
Simple forms of exploration based on action sampling struggle in sparse-reward or long-horizon problems that require complex exploration behaviors \citep{mnih2013playing,sutton2018reinforcement}.
To address these shortcomings, some works model the uncertainty around the dynamics or state of the environment and use this as a learning signal to explore the environment \citep{burda2018exploration,pathakicml17curiosity}.
Other methods introduce temporally extended exploration without explicit uncertainty estimation \citep{osband16,plappert2017parameter,bellemare2016unifying,martin2017count,tang2017exploration}.
Hierarchical RL \citep{sutton1999between,nachum2018near} and options learning \citep{bacon2017option} seek to better handle exploration over long-horizon tasks.
Like these works, \method also focuses on complex exploration strategies but leverages the strengths of LLMs to do so.
Other works use language for hierarchical control, but do so for communication between planner and control policies \citep{belkhale2024rt}, not to improve exploration for solving harder tasks as in \method.

Exploration in RL is especially important for LLM-based policies as they typically operate in complex language action spaces, such as open-ended code generation, and are rewarded for achieving sparse outcomes, such as calling a sequence of tools that accomplishes a goal.
Prior work focusing on non-agentic reasoning tasks suggests that RL for LLM post-training primarily refines existing abilities of the base LLM instead of teaching the LLM to solve new problems that are unsolvable by the base model \citep{yue2025does,wu2025invisible}.
We show that the improved exploration from \method helps exceed these limits in agentic tasks.
Works attempt to maintain high output entropy throughout training; however, these techniques are more designed to help training stability and less to promote exploration \citep{cui2025entropy,wang2025beyond}.
Other works directly train for policy diversity using a ``\passk'' reward during training, where the model is rewarded if any attempt out of $k$ attempts succeeds \citep{chen2025pass,tang2025optimizing}.
These methods also try to prevent policy collapse, but still require at least one of the $k$ attempts to succeed under the base model to obtain a learning signal.
Unlike these works, our method incentivizes exploration in problems where the base model achieves minimal starting \passk.

Other methods introduce exploration-specific techniques to LLM RL training.
\cite{gao2025navigate} uses the random network distillation framework \citep{burda2018exploration} to reward novel token sequences while \cite{cheng2025reasoning} directly rewards the policy entropy.
Our method also incentivizes exploration, but does so through language reasoning, and we demonstrate that it empirically outperforms these added exploration objectives.
Furthermore, these works focus on LLM reasoning for math or reasoning problems with a question-and-answer format, while we focus on agentic settings where an LLM agent takes a series of actions to achieve a goal.
\citet{song2025outcome} adds a UCB exploration reward over the predicted final answer, but we focus on action spaces with an unbounded number of action possibilities.
\citet{li2025jointly} uses a pretrained diversity classifier score as a reward, which we do not have in our multi-step agentic setting.
Other works use the pretrained knowledge of LLMs to provide policy feedback \citep{klissarov2023motif}, which is orthogonal to how \method uses the LLM to reason over strategies for better exploration in sparse reward settings.
Works also use LLMs to guide exploration in other non-LLM policies \citep{hao2025llm,tam2022semantic}, whereas we focus on exploration in a single end-to-end LLM agent.

Other works on post-training LLMs for reasoning also introduce language abstractions or plans of reasoning steps or solutions.
\citet{qu2025rlad} uses distillation to first output a ``reasoning abstraction", similar to a strategy in \method.
However, \method does not use a teacher model and introduces diversity-maximizing techniques in RL such as mixed-temperature sampling and strategy reflection.
\citet{wu2025mode} also requires multiple expert models to generate different modes and focuses on distillation rather than online RL, like in \method.
\citet{chen2025nudging} introduces a ``hint" to guide policy responses, but this hint is generated from the ground truth answer.
Unlike these works, \method operates in multi-step agentic settings where only the desired final outcome is specified, such as passing a set of tests, and the ground truth correct action response is unknown.

\section{Approach}

\subsection{Preliminaries}

We formalize the agentic tasks we focus on as Partially Observable Markov Decision Processes (POMDPs) with goal space $ \mathcal{G}$, observation space $ \mathcal{O}$, action space $ \mathcal{A}$,  environment transition function $\mathcal{T}$, and a reward model $R$. For brevity, we omit other elements of the MDP. In the settings we study, $\mathcal{G}$ is a natural language description of the goal, $ \mathcal{O}$ is a textual or visual input, and $\mathcal{Y}$ is the space of any textual outputs by the LLM. 
The environments we consider involve actions like tool-calling and coding, thus the action space $\mathcal{A} \subset \mathcal{Y}$ is expressed as natural language. 
The reward model $R$ is a sparse reward assigned at the end of the episode for successfully completing the task. 

We consider an LLM policy $\pi : \mathcal{G} \times \mathcal{O} \rightarrow \mathcal{Y} \times \mathcal{A}$ with the goal of maximizing the POMDP cumulative reward. At the episode start, the LLM receives a goal $g \sim \mathcal{G}$ and starting observation $o_1 \in \mathcal{O}$. First, the LLM policy samples an intermediate reasoning trace (chain of thought) $y_1 \sim \pi(\cdot \mid g, o_1)$ and then produces the action $a_1 \sim \pi(\cdot \mid g, o_1, y_1)$ which is executed in the environment and returns the next observation $o_2 = \mathcal{T}(o_1, a_1)$.
This processes repeats to sample a trajectory $o_1, y_1, a_1, o_2, y_2, a_2, \dots, a_T, o_T, r_T$ where $r_T$ is the outcome reward. 
The objective is to train the LLM policy with RL to maximize the expectation of rewards over $\mathcal{G}$.

\subsection{\Method}

A key challenge in RL is learning from sparse reward functions that only score the binary success of achieving the goal.
LLM agents operating in complex environments and language action spaces are unlikely to stumble onto the goal by chance, thus the starting LLM must already have the capability to intelligently search to find the goal.
Prior work shows that this reliance on the base model's sampling prevents discovering truly new solutions \citep{yue2025does,wu2025invisible,sun2025delta}.

We thus introduce \Method (\method) to incentivize the LLM policy to explore and discover solutions to tasks unsolvable by the starting policy.
The key idea of \method is for the policy to produce high-level language strategies before actions, and then enforce diverse strategy sampling to drive exploration.
\method does not require ground truth solutions, privileged information or access to a stronger LLM, and instead modifies the LLM sampling process during RL training.
Specifically, \method makes three modifications over standard RL training: (1) strategy prompting, (2) mixed-temperature sampling, and (3) strategy reflection. 

\textbf{Strategy Prompting}:
For goal $g \in \mathcal{G}$ and observation $o_t$ at episode step $t$, instead of sampling an intermediate reasoning trace $y_t \sim \pi(\cdot \mid g, o_t)$, we first sample a strategy $s_t \sim S_\pi(\cdot | g, o_t)$ from ``strategy sampling distribution" $S_{\pi}$. 
For clarity, we subsume the goal and observation $(g, o_t)$ into just the observation $o_t$.
The intermediate reasoning trace and then action is generated with this strategy, meaning the full output distribution is $\pi(a_t | y_t, s_t, o_t) \pi(y_t | s_t, o_t) S_\pi(s_t | o_t)$.
The strategy is a concise language description that maps to a specific and non-overlapping distribution of actions.
As we will empirically demonstrate, $S_\pi$ induces actions that explore the environment more efficiently than general intermediate reasoning or direct action sampling. 
The strategy is different from the other intermediate chain-of-thought text or action tokens in that it is sampled from the different distribution $S_\pi$, which uses a specific prompt and two additional techniques for generating diverse strategies described later in this section.

Specifically, for a given observation, \method modifies the prompt to include text like: \emph{First, give a strategy of what action to take after `Strategy:' then generate the action after `Action:'}.
Simply modifying the prompt to produce a strategy before the action is not particularly novel, with prior works also showing the benefits of first producing a strategy-like output before the response \citep{qu2025rlad,chen2025nudging}.
The key difference in \method is how the strategy is leveraged for exploration in RL with the next two techniques.

\textbf{Mixed-Temperature Sampling}: 
This generates strategies at a higher token sampling temperature than typical for the model inference.
This higher temperature only affects the tokens sampled from the strategy distribution $S_\pi$ and not the remaining LLM outputs from $\pi$.
This means that the model response is decoded with a mixed-temperature, where the initial strategy is decoded with a high temperature and the remaining tokens are decoded with a comparatively lower temperature.
The insight is that diverse strategies will more effectively explore the environment than directly sampling more diverse actions. 
Sampling actions with a high temperature can result in actions that are only superficially distinct. 
For example, in UI control, high-temperature action sampling can result in the agent tapping at different coordinates on the same button. 
This added noise could hinder RL if, for example, the UI agent's noisy tap locations sometimes click off the intended button. 
We empirically find it is easier for the model to sample different strategies, and then condition the lower-temperature action generation on these strategies.
To explore different strategies, \method produces $K$ parallel strategies per task, which has no added generation overhead over the group-based RL algorithms typically used in LLM RL training, such as GRPO \citep{deepseek-math} or RLOO \citep{ahmadian2024back}, which require $K$ parallel responses per task for advantage estimation.

\textbf{Strategy Reflection}: During RL training, \method also improves the diversity of strategy generation based on the environment feedback.
With some probability, \method does ``Negative Strategy Reflection''.
If a strategy fails (meaning the outcome reward $r_T=0$), then we condition a subsequent rollout in the same task on the failed strategy with a negative reflection prompt instructing the policy to critique the failed strategy and to generate a new strategy that addresses the flaws of the failed strategy. 
If the environment also provides episode-level textual feedback in addition to the scalar reward $r_T$, we also include this with the strategy.
For example, in the case of coding, the output of a failed program provides textual output of failed tests or runtime errors.
Likewise, \method also incorporates ``Positive Strategy Reflection'' where, with some probability, strategy generation is conditioned on a strategy from the same task that ended in success. 
The prompt instructs the agent to produce a new strategy inspired by the successful example in the same task.
We find that the negative reflection enables the policy to solve new episodes beyond parallel strategy sampling. 
Positive reflection results in better learning efficiency as the agent is able to better leverage successful examples and maintain higher output entropy with multiple successful strategies solving the same task.
Since the reflection process happens during RL, the strategies are from old versions of the policy. 
This further boosts strategy diversity over the strategies purely generated from a single policy instance.

We detail all the prompts, including the strategy and reflection prompts in \Cref{sec:prompts}.
All prompts are short at mostly two sentences and are consistent between environments with slightly different language in how actions are referred to depending on the environment and LLM being trained.
Environments are multi-step, so we generate a per-action strategy at every step. 
In strategy reflection, we condition the reflection process on the entire sequence of strategies from the previous episode.
We only use strategy reflection at training time, and evaluate the \method trained policy in a standard setting.
Our evaluation domains have on the order of 10's of steps per episode, which keeps this extra context manageable.
Future work extending \method to environments with more steps could summarize long sequences of per-step strategies into a more compact meta-strategy to save the cost of the extra context.

\subsection{RL Details}
\label{sec:rl-details}

We use \method with GRPO \citep{deepseek-math}. 
However, since \method only alters the sampling distribution via mixed-temperature sampling and policy conditioning via strategy reflection, it is compatible with any online RL algorithm.
Between inference and training, our implementation uses a consistent per-token temperature, which varies between the strategy and remaining tokens, to ensure consistency in the policy distribution between the two stages.
We also use the GRPO modifications proposed by DAPO~\citep{yu2025dapo}, except without the clip higher probability ratio, which we found was broadly harmful across all our experimental domains and baselines.
We detail all method details, prompts, and hyperparameters in \Cref{sec:method-details}.

\begin{figure*}[!t]
  \centering
  \begin{subfigure}[b]{0.50\textwidth}
    \includegraphics[width=\textwidth]{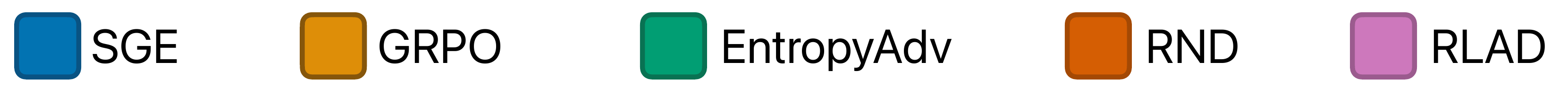}
  \end{subfigure}
  \par\medskip %
  \begin{subfigure}[b]{0.24\textwidth}
    \includegraphics[width=\textwidth]{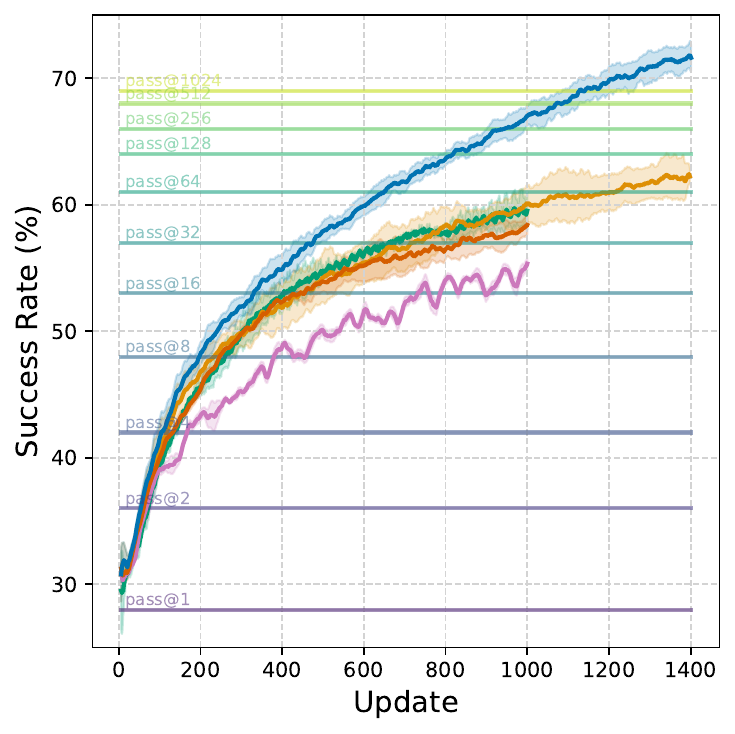}
    \caption{Coding}
    \label{fig:sub1}
  \end{subfigure}
  \begin{subfigure}[b]{0.24\textwidth}
    \includegraphics[width=\textwidth]{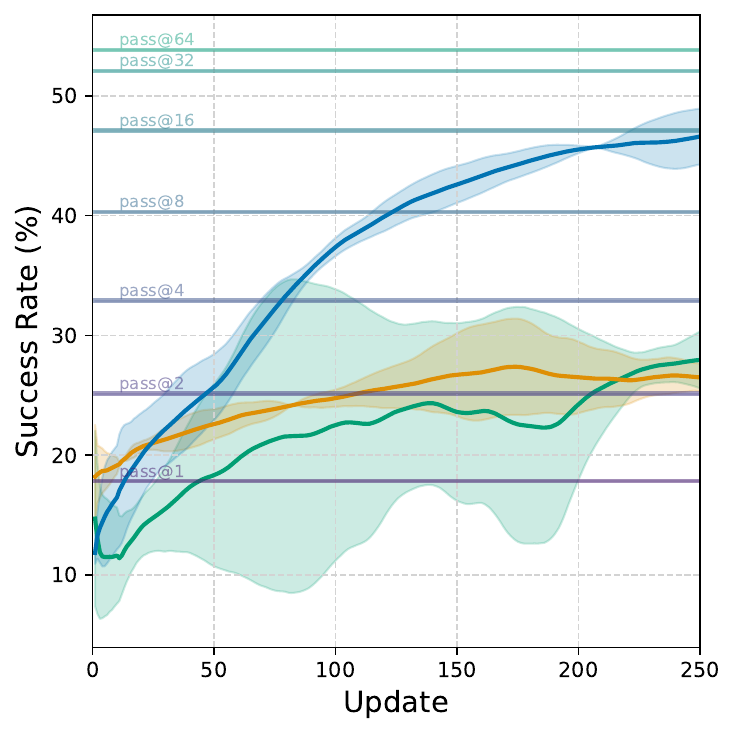}
    \caption{AndroidWorld}
    \label{fig:sub3}
  \end{subfigure}
  \begin{subfigure}[b]{0.24\textwidth}
    \includegraphics[width=\textwidth]{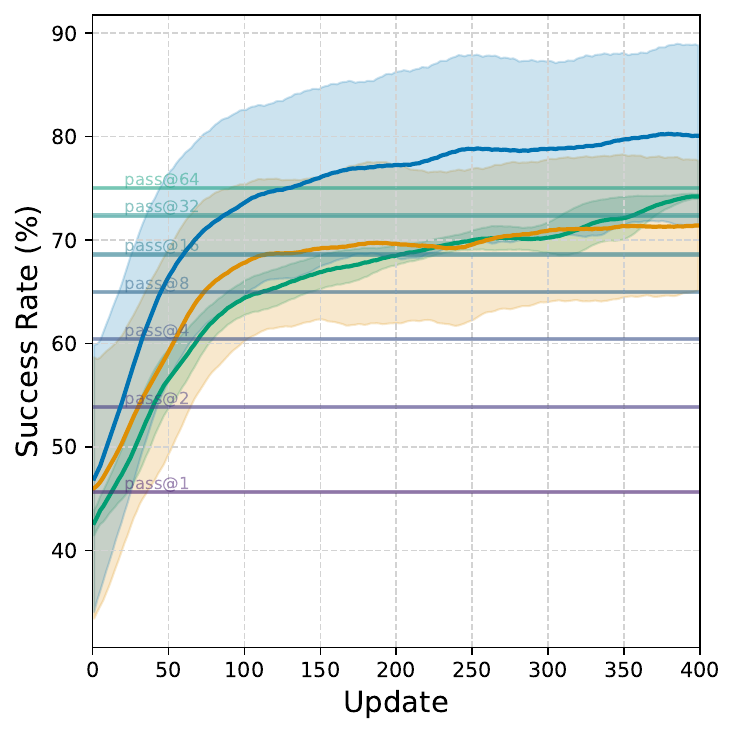}
    \caption{LangR}
    \label{fig:sub2}
  \end{subfigure}
  \begin{subfigure}[b]{0.24\textwidth}
    \centering
    \includegraphics[width=\textwidth]{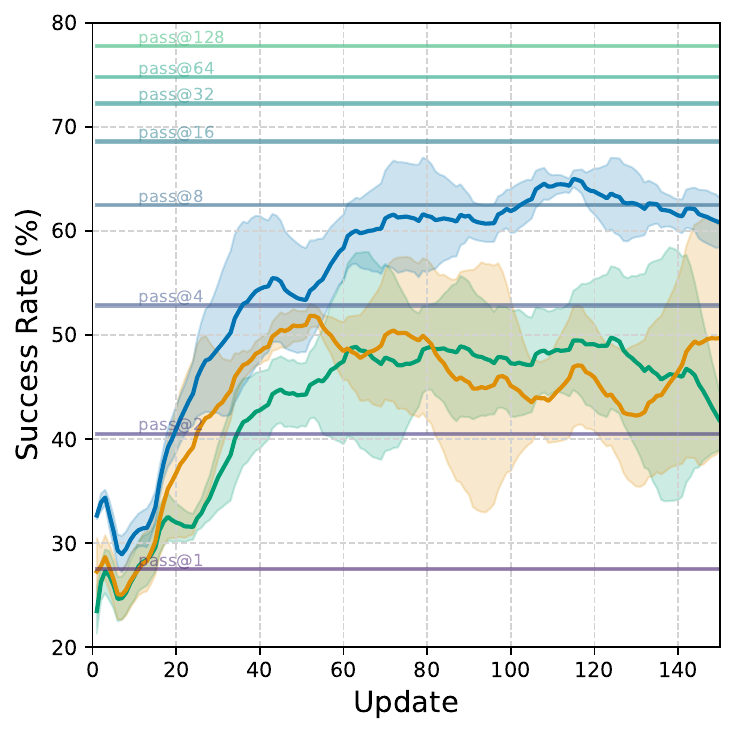}
    \caption{AppWorld}
    \label{fig:sub4}
  \end{subfigure}
  \caption{
    RL training curves for \method and baselines in each environment.
    Training curves show the average pass@1 rate versus the number of RL updates for the training tasks across 3 random seeds in training runs, with the shaded area representing the standard deviation of the pass@1 rate across the seeds.
    The horizontal lines indicate the base model \passk rate with lighter shades meaning a higher $k$.
    In the Coding environment, \method and GRPO are trained longer than the other approaches to show better converged performance.
    RND and RLAD baselines are also only shown for this environment since they underperform the GRPO and EntropyAdv.
  }
  \label{fig:main-results}
\end{figure*}

\section{Experiments}
\label{sec:experiments} 

\subsection{Environments}
\label{sec:envs} 
We show results across four diverse agentic domains to demonstrate the broad applicability of \method.
\begin{itemize}[itemsep=3pt,topsep=0pt,parsep=0pt,partopsep=0pt,parsep=0pt,leftmargin=*]
  \item AndroidWorld \citep{rawles2024androidworld} is a phone UI control environment where the agent operates from RGB pixel inputs and outputs low-level UI interactions, including tapping pixel coordinates, swiping, or typing text.
    We use 26 of the tasks for training and evaluate on 30 test tasks consisting of non-visual question answering tasks in the overall AndroidWorld task set.
    Both the train and test sets span 15 apps, with the test set including tasks on 4 apps unseen in the training set.

    \item Language Rearrangement (LangR) \citep{szot2023large}: An embodied household agent is instructed to rearrange objects in a home environment to complete a language instruction, such as ``Find and put all the apples in the fridge''. We use a textual environment observation, which is a partially observable textual representation of the robot's vicinity, including the object the robot is currently holding, and which objects and receptacles are in front of the robot. The action space includes high-level skills such as picking objects and navigating to receptacles. We use the standard train split and evaluate generalization to unseen houses.

    \item Coding \citep{xia2025leetcodedataset}: Given a coding problem description, the agent must output Python code to pass a set of unit tests.
      We use only the ``hard'' category of problems from the dataset, giving 606 tasks from the train split and 228 tasks from the test split.
      We make this environment multi-turn by providing the output of the executed program as an observation for the policy to produce updated Python code in another attempt.

    \item AppWorld \citep{trivedi2024appworld}: A multi-step tool-calling benchmark in a simulated app ecosystem across 9 applications and 457 APIs.
      We only train and test on the ``Easy'' categorization of problems, which are still challenging for the LLMs we finetune.
      This filtering results in 36 training problems, and we evaluate generalization to unseen tasks with the 57 tasks from the ``Test Normal'' split.
\end{itemize}
 We train separate models for each environment. See \Cref{sec:env-details} for full details on all of the environments.

\subsection{Baselines}
\label{sec:baselines} 

We compare \method against the following approaches for LLM RL training, with a focus on baselines that are intended to improve exploration.
\begin{itemize}[itemsep=3pt,topsep=0pt,parsep=0pt,partopsep=0pt,parsep=0pt,leftmargin=*]
  \item \textbf{GRPO}: Regular GRPO with the standard policy gradient objective. 

  \item \textbf{Entropy Advantage (EntropyAdv)}~\citep{cheng2025reasoning}: Modifies GRPO to incentivize exploration by adding the policy entropy to the advantage for every token. 
    This approach was demonstrated to be more effective for LLM RL exploration than standard entropy regularization that adds an entropy loss term to the RL objective \citep{schulman2017proximal}.

  \item \textbf{Random Network Distillation (RND)}~\citep{gao2025navigate}: Adds an exploration reward for producing novel actions using random network distillation (RND) \citep{burda2018exploration}. RND trains a network to predict the output of a randomly initialized frozen target network based on the final LLM activation for each action. The prediction error is provided as an exploration reward when the agent fails to achieve the goal, encouraging the model to explore novel output sequences.

  \item \textbf{RL with Abstraction Discovery (RLAD)} \citep{qu2025rlad}: This work first outputs an abstraction for the problem, akin to the strategy in \method, and then conditions the response on the abstraction. RLAD trains the abstraction and solution generator by augmenting training with a KL divergence penalty and sometimes dropping out the strategy. We use the same LLM weights for both roles of abstraction and solution generator. RLAD serves as a baseline also addressing explicit strategy generation.

  \item \textbf{Base model pass@k}: This zero-shot evaluates the base model with $k$ independent attempts per task and is the percentage of the time that any of the attempts are successful as judged by the ground truth task verifier. This shows how methods improve on the existing capabilities of the base model.

\end{itemize}
We find prompting the policy to produce a strategy before the action to be marginally useful for even non-\method baselines; therefore, all baselines use this improved prompt for a fair comparison (see \Cref{sec:prompt-impact} for more details).
Thus, the difference between \method and the standard GRPO baseline is that \method uses the mixed-temperature sampling and strategy reflection.
All methods, \method included, are trained with the GRPO.
For LangR and Coding, we finetune the Qwen3-4B-Instruct model \citep{yang2025qwen3}.
For AppWorld, we finetune the Qwen3-8B reasoning model, as we found the larger size was necessary for stable GRPO training.
For AndroidWorld, which has visual observations, we finetune the Qwen2.5-VL-3B model \citep{bai2025qwen25vltechnicalreport}. 
We only run the RND and RLAD baselines on Coding since they underperform the GRPO and EntropyAdv baselines.
All hyperparameters and full implementation details are provided in \Cref{sec:hyperparams}.

Note that since we operate in an agentic setup where an agent needs to take actions to achieve a goal, we cannot apply some of the methods from prior works that are specific to single-step reasoning question answer domains. 
We cannot apply the UCB exploration method from \citet{song2025outcome} since there is an unbounded number of action trajectories that can lead to the goal. 
The agentic environments we evaluate in also do not contain reference solutions, so we cannot compare to the method from \citet{chen2025nudging}, which evaluates in the math domain and assumes access to the problem answers. \citet{li2025jointly} assumes access to a pretrained diversity classifier, which is difficult to apply to strategy and action sequences.

\begin{figure*}[t!]
  \centering
  \begin{subfigure}[b]{0.24\textwidth}
    \includegraphics[width=\textwidth]{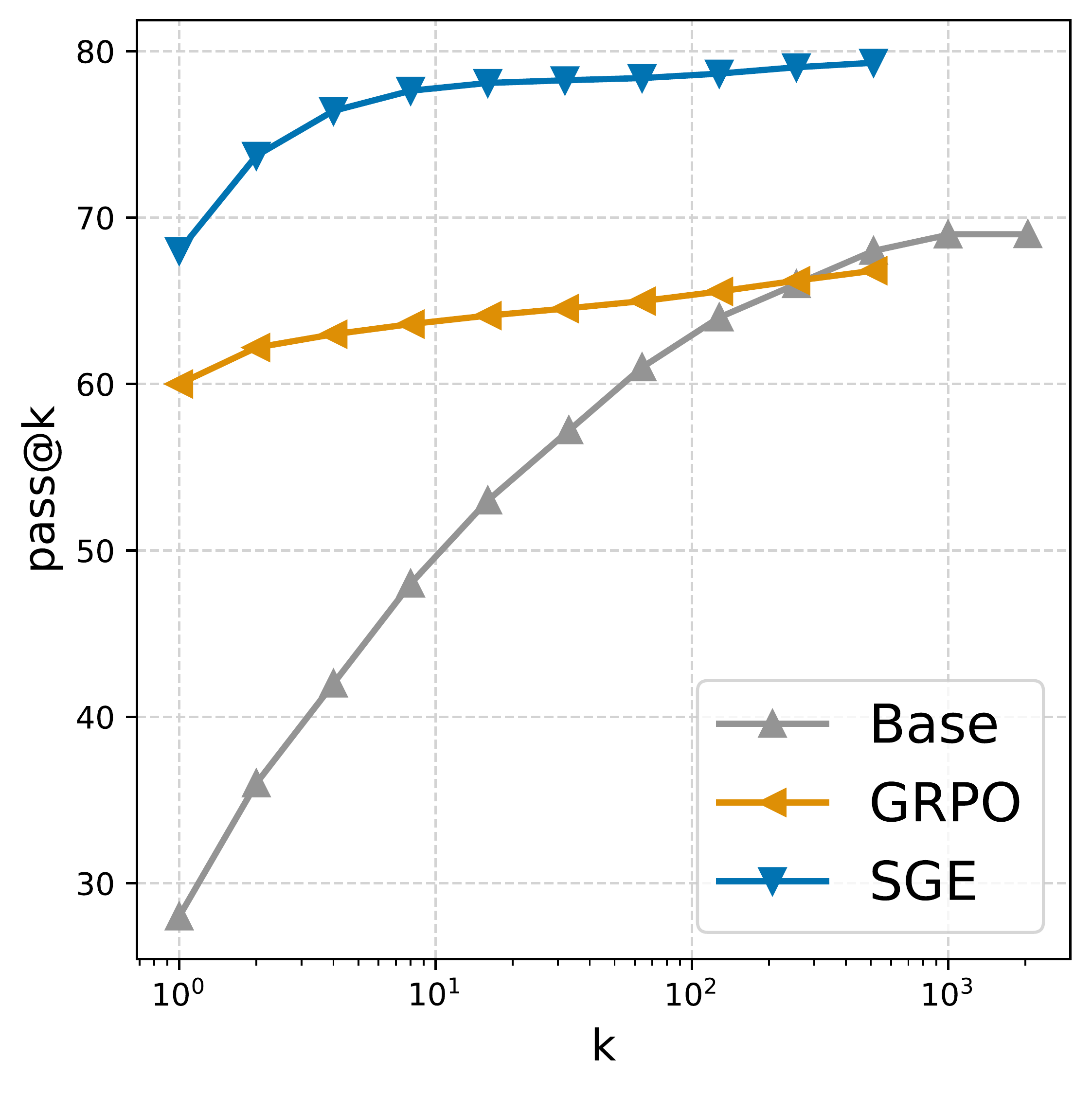}
    \caption{Coding}
  \end{subfigure}
  \begin{subfigure}[b]{0.24\textwidth}
    \includegraphics[width=\textwidth]{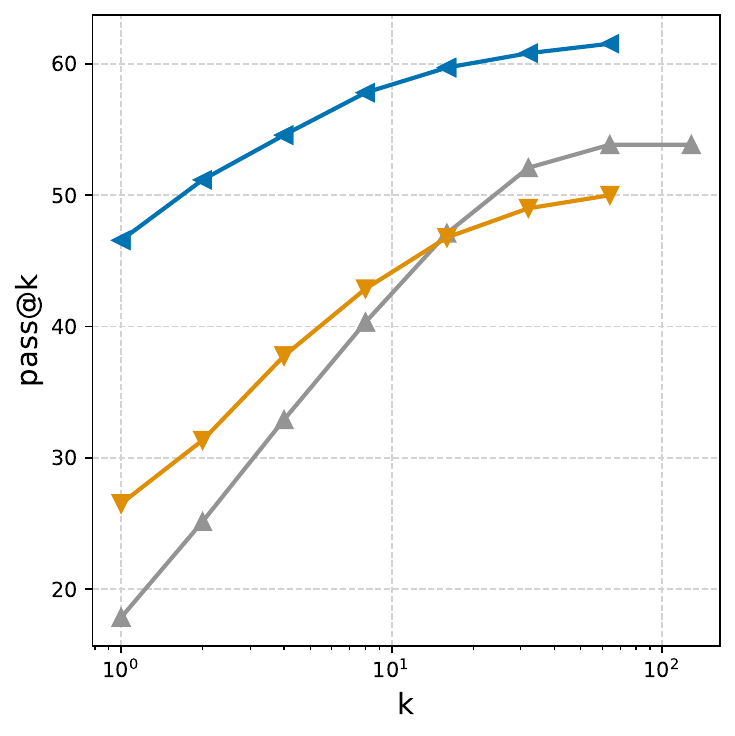}
    \caption{AndroidWorld}
  \end{subfigure}
  \begin{subfigure}[b]{0.24\textwidth}
    \includegraphics[width=\textwidth]{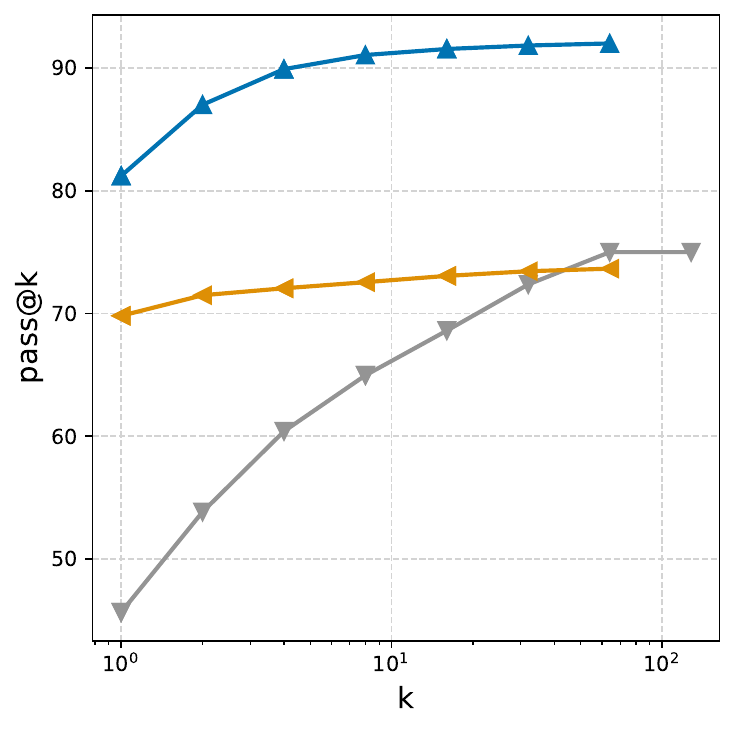}
    \caption{LangR}
  \end{subfigure}
  \begin{subfigure}[b]{0.24\textwidth}
    \includegraphics[width=\textwidth]{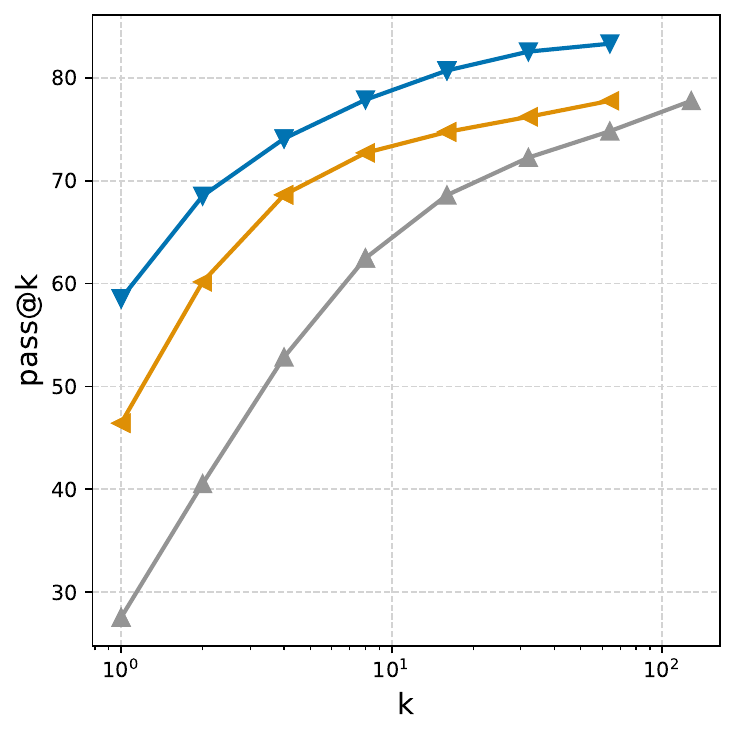}
    \caption{AppWorld}
  \end{subfigure}
  \caption{Comparing the \passk of the final policies trained by \method and GRPO and the starting base model across the environments on the training tasks. We compute the \passk for $k$ in powers of 2, and the x-axis is on a logarithmic scale. The ``Base'' line shows that the \passk of the base model eventually plateaus for large enough $k$. GRPO raises the ``Base'' curve, enabling the pass@1 to closely match the \passk for the highest $k$. On the other hand, \method exceeds this \passk ceiling of the base model and enables solving new problems that are unsolvable by the base model. We report averages across all tasks in the training set.
}
  \label{fig:train-passk}
\end{figure*}

\subsection{Empirical Comparison to Baselines}
\label{sec:exps-prior-work} 

\textbf{RL training results:}
We compare \method RL training to baselines in \Cref{fig:main-results}, which demonstrates \method achieves higher RL training performance.
Specifically, we plot the average pass@1 rate, which is the percentage of the time the agent succeeds in solving the problem on the train set, along with the standard error across 3 random training seeds in RL training.
The overlaid horizontal lines indicate the \passk of the base model for $k$ values up to where the \passk does not change when doubling $k$.
This means the base LLM cannot solve any more tasks, even with twice as many attempts.
The plots demonstrate that \method achieves higher final training performance than baselines across all environments, with on average \method achieving 27\% higher relative final success than the next best baseline for that environment.
Notably, baselines are unable to surpass the maximum \passk performance of the base model, confirming the findings from works such as \citet{yue2025does} for agentic settings.
However, in Coding and LangR, \method surpasses the maximum \passk by on average 11\% relative performance, demonstrating that the exploration from \method enables the model to learn new behaviors not demonstrated by the base model.

\method also outperforms the EntropyAdv and RND baselines, which also focus on better exploration in LLM RL. 
We find that while EntropyAdv and RND increase policy token output entropy in RL training, unlike \method, they cannot explore to exceed the maximum \passk.
These methods were developed in the context of question answering in math reasoning problems, while we focus on agentic exploration, where diversity over outcomes is more important than diversity over token outputs.
Different output tokens may correspond to actions that produce similar effects, for example, the agent may make only minor variations to the syntax of the code in the Coding environment or tap the same button at different positions in AndroidWorld.
\method also outperforms RLAD, another method that leverages strategies.
RLAD is hampered by slow training, potentially due to the KL divergence penalty required in the method.
RLAD also only showed results for non-agentic question answering problems for the smaller Qwen3-1.7B model.
\method better leverages strategy generation for exploration in agentic RL.

Next, in \Cref{fig:train-passk} we plot the \passk curve of the trained policy across \method and baselines.
Regular GRPO training ``flattens'' the \passk curve, raising the pass@1 value with RL to closer match the pass@k value.
On the other hand, \method benefits from increased test time scaling by increasing in performance with increased $k$ to consistently surpass the base model.
This indicates that \method enables the policy to solve new tasks that are not solvable by the base model, even for very large $k$, where the \passk has plateaued for the base LLM.
This also occurs in AndroidWorld and AppWorld, where the RL pass@1 does not surpass the base model max \passk.
\begin{wraptable}{r}{0.5\textwidth}
\vspace{0.4em}
\centering
\resizebox{\linewidth}{!}{%
\begin{tabular}{lccc}
\toprule
 & \textbf{Zero-Shot} & \textbf{GRPO} & \textbf{SGE} \\
\midrule
Coding         & 13.5 & 22.0 {\scriptsize $\pm$ 0.3} & \textbf{29.2} {\scriptsize $\pm$ 0.7} \\
AndroidWorld  & 16.7    & 21.9 {\scriptsize $\pm$ 0.3}  & \textbf{36.7} {\scriptsize $\pm$ 1.3} \\
LangR         & 42.6 & 46.0 {\scriptsize $\pm$ 0.4}& \textbf{60.8} {\scriptsize $\pm$ 0.6}\\
AppWorld      & 47.8 & 49.3 {\scriptsize $\pm$ 5.0} & \textbf{66.6} {\scriptsize $\pm$ 2.9} \\
\bottomrule
\end{tabular}
}
\caption{
  Test evaluation on unseen tasks of \method versus baselines after RL training. 
  Numbers are average and standard error pass@1 across the 3 random seeds on unseen tasks not included in RL training. 
  All coding checkpoints are evaluated after 1k updates.
  Zero-Shot refers to evaluating the base-LLM directly on the test set without any RL training. 
  RL improves generalization, and \method further boosts it.
}
\label{tab:gen}
\vspace{-1em}
\end{wraptable}

\textbf{Generalization to unseen tasks}: 
Next, in \Cref{tab:gen}, we evaluate the generalization of the RL-trained policies to unseen task instances using the test splits earlier described in \Cref{sec:envs}.
The results in \Cref{tab:gen} show that \method trains policies that generalize to new problems better than baselines across the four considered environments.
This shows that \method not only explores the train environments better but also learns generalizable behaviors that transfer to unseen tasks, indicating that the agent is able to utilize its ability to solve harder problems during training time on the unseen test tasks.

\begin{figure}[t!]
  \centering
  \begin{subfigure}[b]{0.32\textwidth}
    \includegraphics[width=\textwidth]{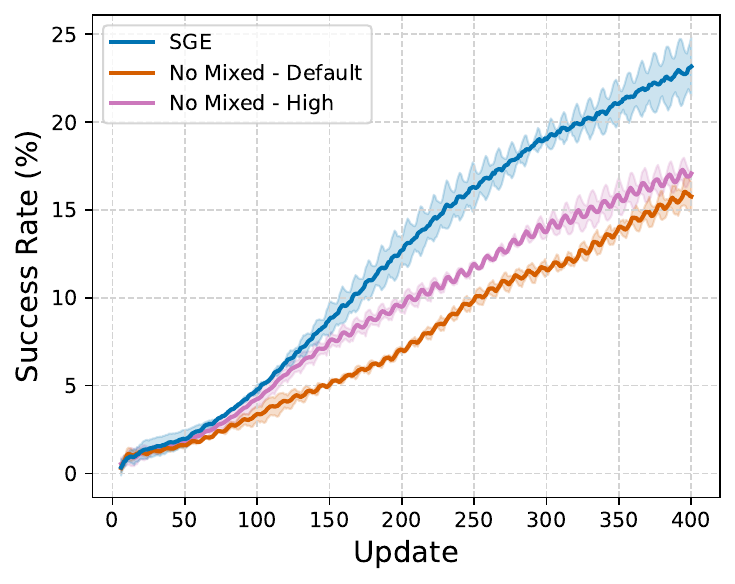}
    \caption{Mixed-Temp Ablation}
    \label{fig:sampling-abl} 
  \end{subfigure}
  \begin{subfigure}[b]{0.32\textwidth}
    \includegraphics[width=\textwidth]{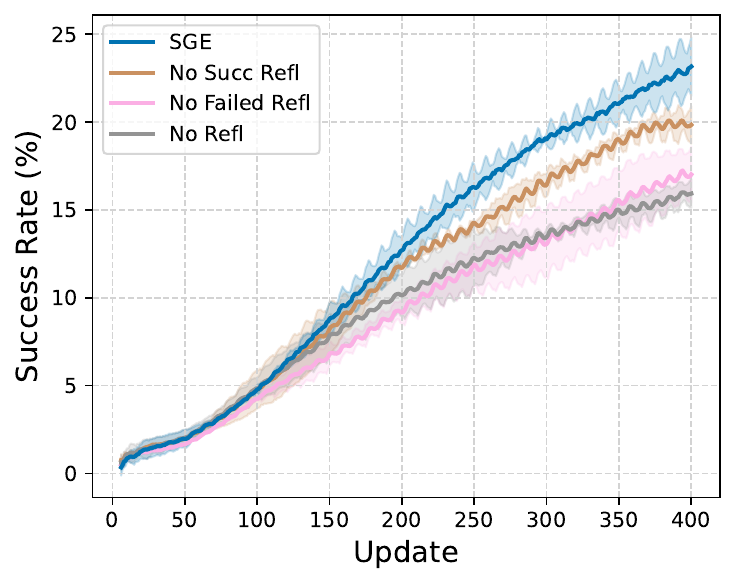}
    \caption{Reflection Ablation}
    \label{fig:refl-abl} 
  \end{subfigure}
  \begin{subfigure}[b]{0.32\textwidth}
    \includegraphics[width=\textwidth]{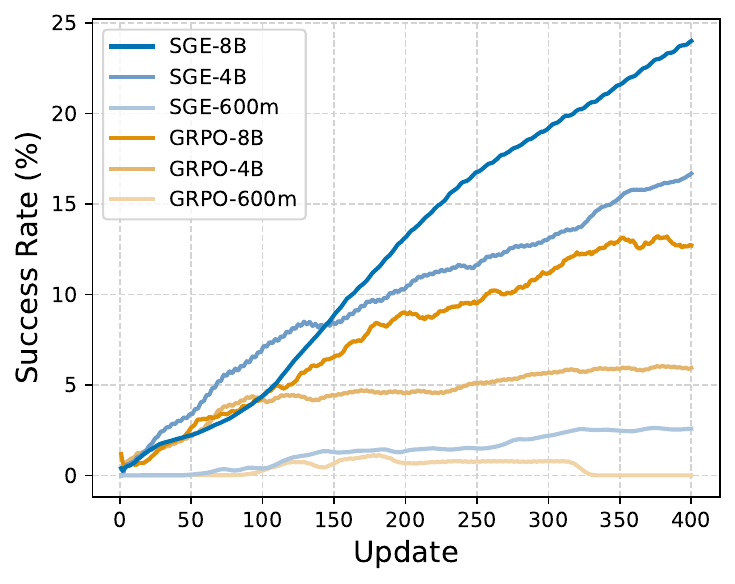}
    \caption{Scaling Results}
    \label{fig:scaling}
  \end{subfigure}
  \caption{
    Left: we ablate the parallel strategy sampling by comparing to sampling all tokens with the uniformly high temperature used to sample the strategy tokens or the LLM default temperature. Middle we ablate the sequential strategy reflection by removing \method's ability to reflect on successful strategies, negative strategies, or any strategy.
    Right: Effect of scaling Qwen3 base model parameter count on RL training with and without \method.
    Ablation results are over 3 random seeds, and scaling results are over 1 random seed.
  }
  \label{fig:strat-analysis}
\end{figure}

\subsection{Analysis}
This section further analyzes \method in a setup that focuses on difficult tasks where the base model struggles.
Specifically, we filter the hard category of the coding environment from \Cref{sec:exps-prior-work} to only the problems where pass@256 of the Qwen3-4B model is $0\%$, which leaves $144$ problems.
These difficult problems test the limit of the model's ability to explore and find creative solutions.
For this section, we finetune the Qwen3-8B model as it has more efficient RL training on these difficult problems for running analyses.

\textbf{\method component ablation}: First, we analyze the contribution of the \method components in \Cref{fig:strat-analysis}.
In \Cref{fig:sampling-abl}, we remove the mixed-temperature sampling from \method and instead sample all tokens with the default action temperature (``No Mixed - Default'') or the high temperature used to sample the strategy (``No Mixed - High'').
\Cref{fig:sampling-abl} shows that either using the default or higher temperature results in worse performance than the mixed-temperature sampling.
\Cref{fig:refl-abl} shows that the failed strategies are more important in the reflection process than the successful strategies.
Both types of reflection help the overall \method performance.
Overall, all components of \method are important for the best performance.

\begin{figure}[t!]
  \centering
  \begin{subfigure}[b]{0.35\textwidth}
    \includegraphics[width=\linewidth]{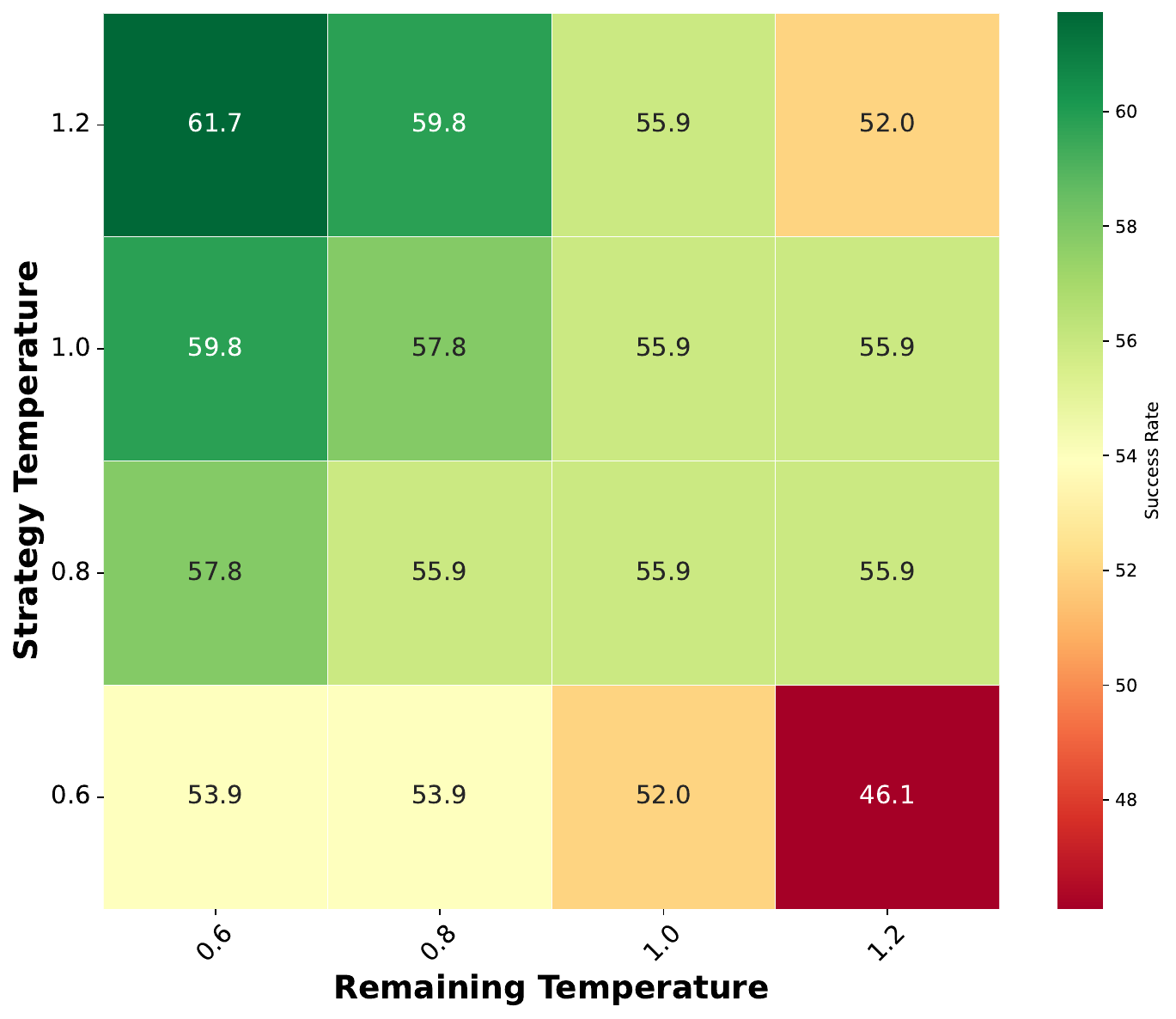}
    \caption{Pass@16 vs. Temperature}
    \label{fig:temp-analysis}
  \end{subfigure}
  \begin{subfigure}[b]{0.6\textwidth}
    \includegraphics[width=\textwidth]{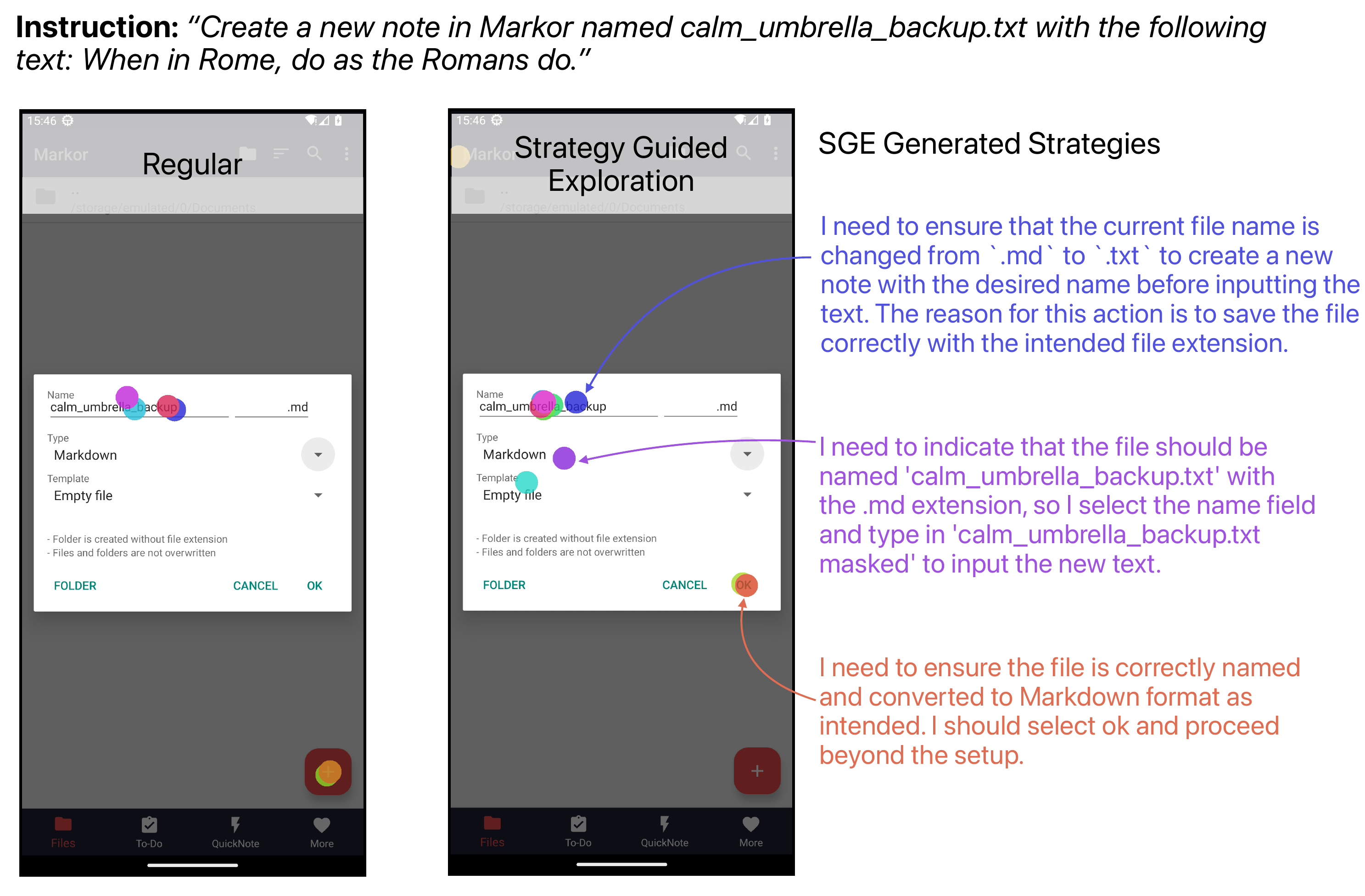}
    \caption{Qualitative Example of \method in AndroidWorld}
  \label{fig:qual-vis}
  \end{subfigure}
  \caption{
    Left: Comparing the effect of different mixed-temperature sampling settings on the zero-shot pass@16 performance in the Coding environment. 
    The high-values in the top-left indicate that a mixed-temperature sampling produces the best results.
    Right: Visualizing actions predicted by \method vs. regular sampling in the \emph{MarkorCreateNote} task in AndroidWorld. 
    16 actions are sampled for the same observation screenshot with the tap locations visualized as colored circles overlaid on the screen (non-tap actions are not visualized).
    To complete the task, the agent must tap the file extension dropdown.
    Unlike with regular sampling, \method is able to sample this correct action.
  }
\end{figure}

\textbf{Effect of mixed-temperature sampling on the base model:} In \Cref{fig:temp-analysis}, we further analyze the effect of mixed-temperature sampling by comparing different temperature values for the strategy and the remaining tokens when sampling from the base model. \Cref{fig:temp-analysis} shows that a high strategy temperature and a relatively lower temperature on the remaining tokens produce the highest pass@16, and thus the strongest exploration results. For RL training, we primarily care about the exploration capabilities of the model; thus, we use this mixed-temperature setting in \method.

\textbf{Effect of model scaling:} In \Cref{fig:scaling}, we analyze the effect of scaling the base LLM parameter count on RL training with or without \method by training with 600M, 4B, and 8B Qwen3 model sizes. 
Expectedly, training greatly improves with larger models, with the positive effects of \method being most prominent at larger model scales.
\method improves training at the 4B and 8B model scales, however, the smallest 600M model achieves close to $0\%$ success and \method only barely helps.
This indicates that \method is still limited by the level of reasoning and planning capability in the LLM, and if the base LLM is not capable of generating or leveraging diverse strategies, \method, like the GRPO baseline, struggles to explore.

\textbf{\method qualitative exploration capabilities:} In \Cref{fig:qual-vis}, we visualize a qualitative example showing the exploration capabilities of \method over standard GRPO training in the AndroidWorld \emph{MarkorCreateNote} task.
In the visualized observation that occurs in the middle of the episode, the agent has already opened the new file dialogue and entered the filename.
Now, the agent must change the file extension from \emph{.md} to \emph{.txt}. 
The QwenVL-2.5-3B model struggles with this step as it frequently tries to change the extension by typing \emph{.txt} into the filename, while only selecting the other file extension in the dropdown actually changes the extension.
The left screenshot of \Cref{fig:qual-vis} shows that the tap locations of the regular policy sampling result in tapping slightly different locations on the filename extension.
However, the right screenshot of \Cref{fig:qual-vis} shows that \method policy sampling taps several locations around the file creation dialogue box, and correctly interacts with the file extension dropdown menu.
Each of these different taps is driven by a different strategy.
While many of the taps are not correct, the correct tap is covered, thus increasing the pass rate for this task as previously empirically demonstrated in \Cref{fig:temp-analysis}.

See \Cref{sec:add-results} for additional experimental results on the impact of the strategy prompt, qualitative examples of the strategy generation, and further analyzing exploration in \method. 

\section{Conclusion and Limitations}
This work introduces \Method (\method), a method for exploration in RL training for LLM agents.
\method uses the textual reasoning abilities of LLMs to produce diverse high-level action strategies and then conditions the action generation on these textual strategies.
These strategies provide an effective space for agents to explore diverse outcomes in the environment.
\method uses mixed-temperature sampling and strategy reflection to improve the diversity of generated strategies and thus achieve better exploration to solve challenging tasks.
We empirically demonstrate that \method outperforms prior RL approaches across a variety of agentic domains in coding, UI control, tool-calling, and embodied AI.

A limitation of \method is that it requires the starting LLM to have sufficient reasoning and planning capacity to generate and leverage the strategies.
We design \method for agentic scenarios where the agent interacts with an external environment, and leave extending \method for other non-agentic RL scenarios, like math reasoning training, for future work.
\method also introduces the additional cost of having to produce a strategy before every step. This limits real-world deployment where response latency is important. 
Future work can investigate how to dynamically predict new strategies only when necessary.

\bibliography{main}
\bibliographystyle{icml2026}

\pagebreak
\clearpage

\appendix
\begin{algorithm*}[t!]
\begin{algorithmic}[1]
\State Hyperparameters: Max buffer size $B$, token sampling temp $\tau$, strategy sampling temp $\tau_s$, good strategy sample probability $p_G$, bad strategy sample probability $p_B$, starting state and goal distribution $\rho$
\State Initialize: policy $\pi_\theta$, success and failure strategy buffer $\mathcal{B}_G, \mathcal{B}_B = \left\{  \right\} $
\For{RL update $i = 1, \dots, N$}
    \State Sample starting State $o_1, g_i \sim \rho$
    \For{Parallel environment rollout for steps $t = 1, \dots, M$ to generate trajectory batch $\left\{ \tau_i \right\}_{i=1}^{K} $}
        \State Set $C_S = \emptyset$, if $\text{Unif}(0,1) < p_B$ then $C_S \sim \mathcal{B}_B$, else if $\text{Unif}(0,1) < p_G$ then $C_S \sim \mathcal{B}_G$
        \State Sample strategy: $s_t \sim \pi_\theta(\cdot | g_i, o_t, C_S; \tau_s)$
        \State Sample remaining tokens: $y_t \sim \pi_\theta(\cdot | s_t, g_i, o_t ; \tau)$
        \State Execute action in environment: $o_t \gets \mathcal{T}(o_t, \text{ExtractAction}(y_t))$
    \EndFor
    \State Store $\tau_i$ in $ \mathcal{B}_G$ if the goal was achieved or $ \mathcal{B}_B$ otherwise.
    \State Do standard GRPO update with collected $\left\{ \tau_{i} \right\}_{i=1}^{K} $.
\EndFor
\end{algorithmic}
\caption{\textsc{\Method} pseudocode.}
\label{alg:sge}
\end{algorithm*}

\section{Implementation Details}
\label{sec:hyperparams} 
\label{sec:method-details}

\subsection{Prompts}
\label{sec:prompts} 

\method requires setting three prompts: (1) the \emph{strategy prompt} which guides the LLM to first generate a strategy before the remaining tokens, (2) the \emph{positive reflection prompt} which conditions the strategy generation on previously executed successful strategies, and (3) the \emph{negative reflection prompt} which conditions the strategy generation on previously executed failed strategies.
The prompts we use for the different environments are largely the same.
Differences are that the action is referred to differently in the environments as ``code'', ``tool call'', or ``action''.
Additionally, the AndroidWorld prompt required formatting the strategy as XML tags to follow the Qwen UI prompt format \citep{bai2025qwen25vltechnicalreport}, whereas the other environments use markdown formatting.
In the reflection prompts, the strategy to reflect on is inserted in \texttt{\{strats\}}.

Coding environment:
\begin{itemize}[itemsep=0pt,topsep=0pt,parsep=0pt,partopsep=0pt,parsep=0pt,leftmargin=*]
  \item Strategy prompt: \texttt{First give a strategy of how to solve the question after ``\#\#\# Strategy''. Then write the code to solve the question based on the strategy and question in ``\#\#\# Code''.}
  \item Positive reflection prompt: \texttt{First give a strategy of how to solve the question after ``\#\#\# Strategy'' inspired by this successful strategy.\{strats\} Then write the code to solve the question based on the strategy and question in ``\#\#\# Code''.}
  \item Negative reflection prompt: \texttt{First, after ``\#\#\# Strategy'' critique the failed strategy and how it can be fixed. Be precise. Then address this critique by writing a better strategy. Make sure the strategy is detailed, and the code is easy to implement from the strategy. \{strats\} Then write the code to solve the question based on the strategy and question in ``\#\#\# Code''.}
\end{itemize}

AndroidWorld environment:
\begin{itemize}[itemsep=0pt,topsep=0pt,parsep=0pt,partopsep=0pt,parsep=0pt,leftmargin=*]
  \item Strategy prompt: \texttt{For each function call, return a strategy in the <strategy></strategy> XML tags and a json object with function name and arguments within <tool\_call></tool\_call> XML tags.}
  \item Positive reflection prompt: \texttt{Here is a previous strategy that was successful at each step: \{strats\}}
  \item Negative reflection prompt: \texttt{Here are the previous sequence of strategies that failed: \{strats\} Your new strategy must be different from this failed strategy sequence and try something new.}
\end{itemize}

LangR environment:
\begin{itemize}[itemsep=0pt,topsep=0pt,parsep=0pt,partopsep=0pt,parsep=0pt,leftmargin=*]
  \item Strategy prompt: \texttt{First give a strategy of how to solve the question after ``\#\#\# Strategy''. Then output the tool call action after ``\#\#\# Action''.}
  \item Positive reflection prompt: \texttt{Here is a previous strategy that was successful at each step: \{strats\} First give a strategy of how to solve the question after ``\#\#\# Strategy'' inspired by this successful strategy. Then generate the tool call in ``\#\#\# Action''.}
  \item Negative reflection prompt: \texttt{Here is a previous strategy that failed: \{strats\} First, after ``\#\#\# Strategy'' critique the failed strategy and how it can be fixed. Be precise. Then address this critique by writing a better strategy. Make sure the strategy is detailed and the code is easy to implement from the strategy. Then generate the tool call in ``\#\#\# Action''.}
\end{itemize}

AppWorld environment:
\begin{itemize}[itemsep=0pt,topsep=0pt,parsep=0pt,partopsep=0pt,parsep=0pt,leftmargin=*]
  \item Strategy prompt: \texttt{Solve this step by step. First think about what your next step should be. Then write the code to execute that step.}
  \item Positive reflection prompt: \texttt{Here is a previous approach that was successful at each step:\textbackslash n\{strats\}\textbackslash n Inspired by this successful approach, first think about what your next step should be. Then write the code to execute that step.}
  \item Negative reflection prompt: \texttt{Here is a previous approach that failed:\textbackslash n\{strats\}\textbackslash nFirst, think and critique the failed approaches and propose how it can be fixed. Be precise. Then address this critique by writing a better approach. Make sure the approach is detailed and the code is easy to implement from the approach. Once you are done thinking, write the code to execute your next step.}
\end{itemize}

In AppWorld, we use the reasoning model and take the entire content in \texttt{\textless think\textgreater...\textless/think\textgreater} as the strategy.
We use the non-reasoning mode for LangR and coding, meaning the model does not first output the \texttt{<think>...</think>} block. 
Instead, we extract the strategy from the ``Strategy" markdown block part of the prompt described in \Cref{sec:prompts}.
Note that even without the explicit ``think" block, the model still produces textual reasoning and a strategy outside of the action.
While we demonstrate \method works both with and without reasoning mode enabled, we use the non-reasoning mode on LangR and Coding as it is more efficient to train since it requires fewer tokens per action, yet it is still sufficient to reason over strategy generation.
Qwen2.5-VL does not have any reasoning model, so we extract the strategy from the prompted \texttt{<strategy>...</strategy>} XML tag.
Note that all methods and baselines use the strategy prompt for all environments, while \method with strategy reflection uses the positive and negative reflection prompts.

\begin{table*}[t]
\centering
\begin{tabular}{lcccc}
\hline
& \textbf{Code} & \textbf{AndroidWorld} & \textbf{LangR} & \textbf{AppWorld}\\
\hline
  Learning Rate &  1e-6 & 1e-6 & 1e-6 & 1e-5 \\
  Group Size & 16 & 6 & 16 & 8 \\
  Global Update Batch Size & 128 & 64  & 128 & 128 \\
  \# Output Tokens / Step & 2048 & 564 & 520 & 768 \\
  Clip Param & 0.2 & 0.2  &  0.2 & 0.2 \\
  Trajectories Collected Per Update & 1024 & 192 & 512 & 128 \\
  Per-Update Epochs & 2 & 2 & 1 & 2 \\
  Token Temperature & 0.7 & 0.7 & 0.6 & 0.7 \\
\hline
  \method Strategy Temperature & 1.2 & 1.2 & 1.0 & 1.2 \\
  \method Strategy Buffer Size & 32 & 32 & 32 & 32 \\
  \method  Failure Strategy Reflection Prob  & 0.25 & 0.25 & 0.25 & 0.25 \\
  \method  Success Strategy Reflection Prob & 0.1 & 0.1 & 0.1 & 0.5 \\
\hline
\end{tabular}
\caption{
RL hyperparameters across domains. Unless otherwise specified, these values are shared between all approaches. \method prefix means the value only applies to \method, and not the baselines. \emph{\# Output Tokens / Step} refers to the total token budget per LLM output for each step in the environment. This multiplies the number of steps in an episode by the total number of tokens produced in an episode. The \emph{Global Update Batch Size} is in terms of the number of steps, not the number of trajectories, which consist of multiple steps.
\emph{Group Size} is the number of trajectories used to estimate the advantage in GRPO. \emph{Trajectories per update} is the number of samples collected in the environment for each update. \emph{Token Temperature} is the temperature used to generate the non-strategy tokens.
}
\label{tab:hyperparams}
\end{table*}

\begin{figure}[t!]
  \centering
  \includegraphics[width=0.4\textwidth]{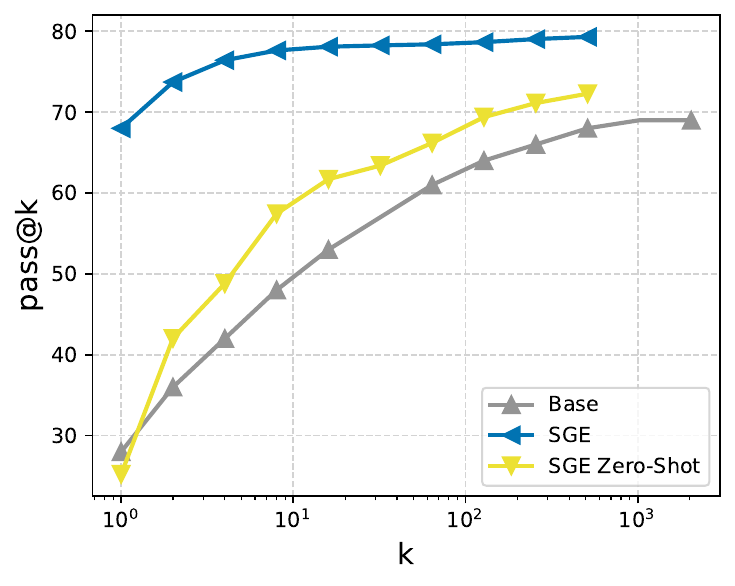}
  \caption{
     We report the impact of \method on the zero-shot performance of the policy in the Coding environment and compare to the base model and SGE RL trained model. We evaluate in the same setup as \Cref{fig:train-passk}.
  }
  \label{fig:sge-passk}
\end{figure}

\subsection{Further \method Details}
See \Cref{alg:sge} for \method pseudocode. Lines 5-10 show how the policy inference during data collection is augmented with the mixed-temperature sampling and strategy reflection. Otherwise, \method follows typical GRPO training. The buffers are implemented as a first-in-first-out buffer to keep the most on-policy data.

\subsection{Hyperparameters}
\label{sec:rl-hypers} 

\Cref{tab:hyperparams} breaks down the hyperparameters for RL training for each environment. 
As \Cref{tab:hyperparams} demonstrates, the values are largely the same between the domains.
Shared RL hyperparameters were first tuned by running regular GRPO in the environment.
Some of the settings relating to the batch size changed in each environment due to the different required context lengths.
The number of output tokens for AndroidWorld and LangR was selected by taking the maximum action length for that environment and adding 512 additional tokens for non-action intermediate outputs for chain-of-thought or strategy generation.
We use a constant learning rate throughout training, and do not employ any learning rate scheduler.
The RL parameters in the top section of \Cref{tab:hyperparams} are shared between all RL methods.
Each RL training job across all environments was run over 16 H100s. The number of updates used to train each method in each environment is displayed in the x-axis of \Cref{fig:main-results}.
For all the experiments, we train all LLM parameters, including the LLM vocabulary embedding and output layers. 
For the Qwen2.5-VL experiments, we freeze the visual encoder module and only train the LLM components.

\subsection{Baseline Details}

\textbf{Entropy Advantage}: This method shapes the per-token advantage term with the token entropy. Specifically, for advantage estimate $A_t$ at output token $o_t$ with output distribution entropy $\mathcal{H}_t$, the advantage term is modified as:
\begin{align*}
  A_t' = A_t + \min \left( \alpha \cdot \mathcal{H}_t, \frac{ |A_t|}{\kappa} \right) 
\end{align*}
Where $\alpha$ is the scaling coefficient and $\kappa$ controls the clipping threshold. As in \citet{cheng2025reasoning}, we set these to $\kappa=2$ and $\alpha=0.4$. Note that the entropy $\mathcal{H}_t$ is detached from the computational graph. This shaping ensures that the entropy does not dominate the advantage term or reverse the sign of the advantage.

\textbf{RL with Abstraction Discovery (RLAD)}: Unlike the other baselines and \method, this approach uses a KL divergence term to the starting policy. We set the KL divergence loss coefficient to be $0.001$. We skip conditioning the solution generation on the strategy $25\%$ of the time in RL training. Note that, unlike \citet{qu2025rlad}, we implement RLAD in a single model where the abstraction and solution generation are performed by a single model.

\textbf{Random Network Distillation (RND)}: RND adapted for LLM post-training is referred to as ``i-MENTOR'' in \citet{gao2025navigate}. To adapt this method to a multi-step decision-making formulation, we use the final token activation for each action in the output sequence for the RND prediction task. 
Specifically, let $y_{t,1}, \dots , y_{t,L}$ be the sequence of output final activations, meaning the final hidden state before the LLM token logit projection head, for step $t$ in the environment, and which are then decoded into action $a_t$.
This hidden state is passed through a randomly initialized MLP, $\overline{f}$, to form a target $\overline{z} = \overline{f}(y_{t,L})$.
This baseline learns an MLP $f$ and is trained to predict the target latent: $\lVert f(y_{t,L}) - \overline{f}(y_{t,L}) \rVert_2^{2}$.
The RND reward is calculated based on
\begin{align*}
  R^{\text{RND}} = \frac{\lVert f(y_{t,L}) - \overline{f}(y_{t,L}) \rVert_2^{2}}{ \text{std} \left( \lVert f(y_{t,L}) - \overline{f}(y_{t,L}) \rVert_2^{2} \right) }
\end{align*}
The RND reward is then assigned to actions from trajectories that receive zero reward.

\section{Additional Results}
\label{sec:add-results} 
\label{sec:prompt-impact} 

\begin{figure}[t!]
  \centering
  \includegraphics[width=0.4\textwidth]{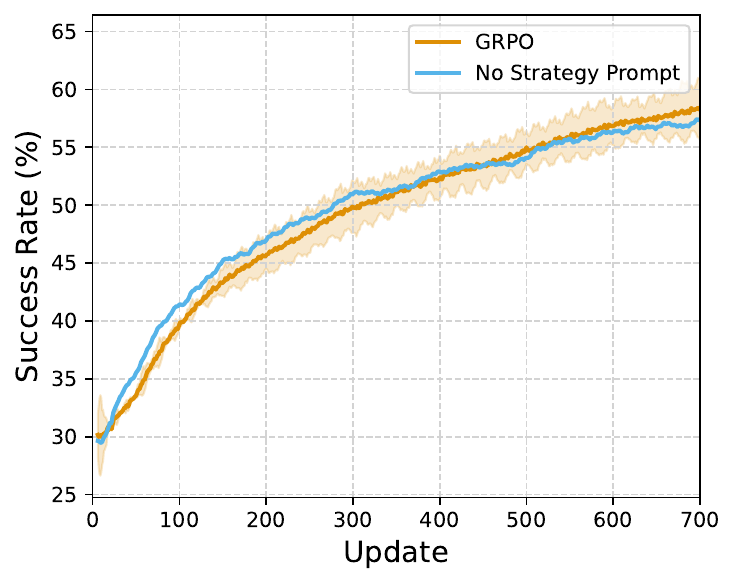}
  \caption{
    Impact of prompting on RL training in Coding environment. Regular GRPO training is with 3 random seeds. Training without the strategy prompt is only with a single random seed.
  }
  \label{fig:lc-prompt}
\end{figure}

\textbf{Impact of strategy prompt}: 
The strategy prompt is a short text asking the LLM to generate a strategy before generating an action.
The exact prompts for each environment are detailed in \Cref{sec:prompts}.
\Cref{fig:lc-prompt} compares standard GRPO training with and without the strategy prompt, demonstrating that the agent performs mostly the same with the strategy prompt and ends with slightly higher performance.
In this result, both approaches were allocated the same token budget for the per-action response.
From these results, all methods in \Cref{sec:experiments} use the strategy prompt for a consistent comparison.

\begin{figure}[t!]
  \centering
  \includegraphics[width=0.4\linewidth]{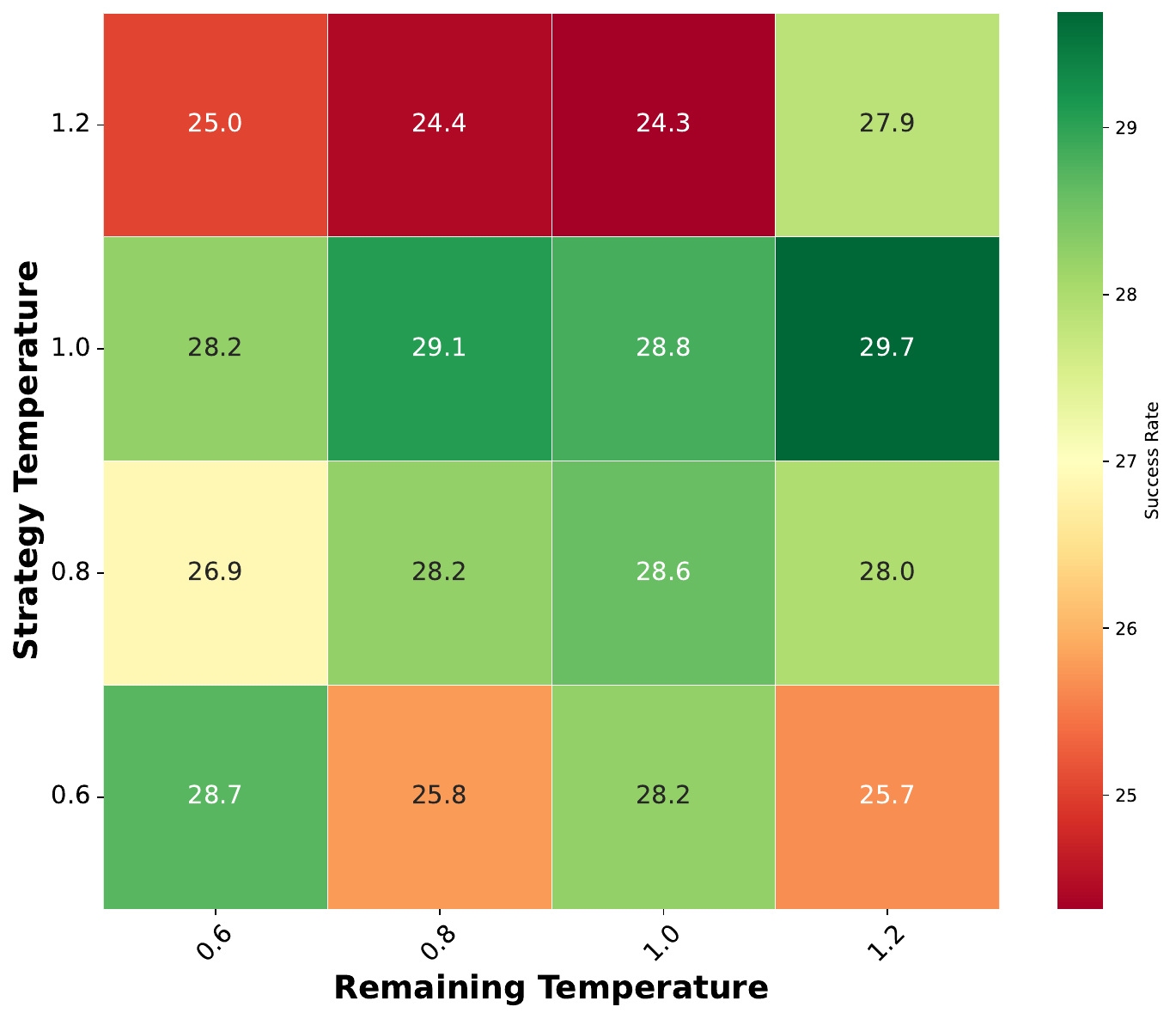}
  \caption{
   Comparing the effect of the sampling temperature on the strategy and remaining tokens in terms of pass@1. Green indicates higher performance. 
  }
  \label{fig:temp-analysis-pass1}
\end{figure}

\textbf{Impact of SGE on zero-shot \passk}: \Cref{fig:sge-passk} shows how the mixed-temperature sampling in \method affects the zero-shot performance of the model.
\Cref{fig:temp-analysis-pass1} breaks down the zero-shot performance across a variety of temperature settings in terms of the pass@1 performance.
The figure demonstrates that the mixed-temperature negatively impacts the pass@1. 
However, this lower pass@1 comes with improved pass@16, as earlier shown in \Cref{fig:temp-analysis}, which ultimately leads to better RL.

\textbf{Qualitative strategy analysis:} We provide qualitative examples of the strategies \method produces during RL training in \Cref{fig:strategy_example1} and \Cref{fig:strategy_example2} for the coding environment. \Cref{fig:strategy_example1} shows an example of how \method generates diverse strategies for the same problem. \Cref{fig:strategy_example2} shows an example of how \method uses negative strategy reflection to improve the diversity of generated strategies and achieve successful outcomes in the environment.

\begin{figure}[t!]
  \centering
  \begin{subfigure}[b]{0.4\textwidth}
    \includegraphics[width=\textwidth]{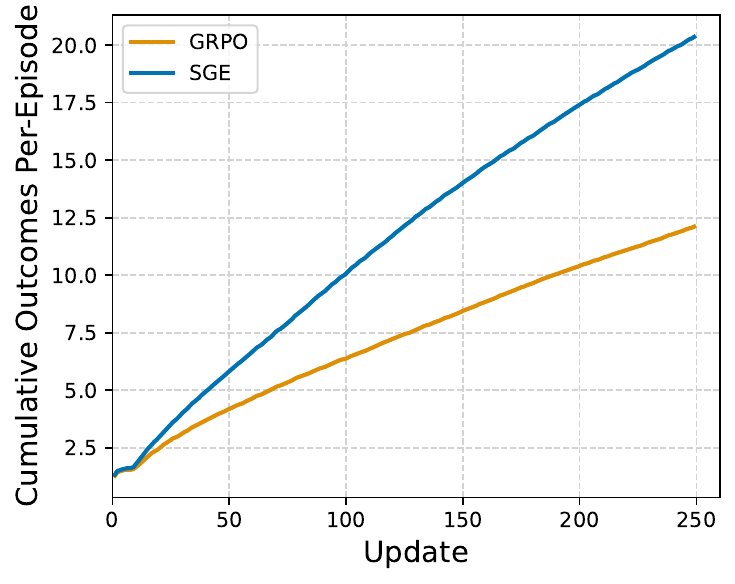}
  \end{subfigure}
  \caption{
    The cumulative unique program outcomes per task in the coding environment during RL training, where each distinct set of tests that passes is considered a distinct outcome. \method achieves more outcomes faster during RL training compared to standard GRPO, demonstrating its effectiveness in exploration.
  }
  \label{fig:expl-outcomes}
\end{figure}

\textbf{State visitation exploration analysis:} We further quantify the exploration capabilities of \method by analyzing state-visitation of the method and regular GRPO training throughout the course of RL in the coding environment.
We do so by measuring the number of distinct outcomes encountered per task during RL training. For the coding environment, we count each distinct set of tests that fails as a distinct outcome. 
This means we measure the functionality of the program rather than the code itself.
\Cref{fig:expl-outcomes} shows that \method encounters more diverse outcomes throughout RL training than regular GRPO.
This means that the programs produced by \method better explore different functionalities in the Python execution environment.
This exploratory behavior helps \method reach the sparse positive outcome more frequently during RL training.

\begin{figure}[H]
\centering
\begin{rawpanel}
\begin{lstlisting}[style=rawtext]
### Question
You are given an integer array nums. The uniqueness array of nums is the sorted array that contains the number of distinct elements of all the subarrays of nums. In other words, it is a sorted array consisting of distinct(nums[i..j]), for all 0 <= i <= j < nums.length. Return the median of the uniqueness array of nums.

<Remainder of problem setup...>
First give a strategy of how to solve the question after "### Strategy". Then write the code to solve the question based on the strategy and question in "### Code".

<Strategy 1>
### Strategy

The problem requires finding the median of the "uniqueness array" of an array nums, where the uniqueness array contains the number of distinct elements in all possible subarrays.

Let's break down the approach:

1. **Key Insight**:
   - The total number of subarrays is n*(n+1)/2 for an array of length n.
   - The uniqueness array will contain values from 1 (single elements) up to at most n (entire array).
   - Since we only need the median (not the full array), we can binary search the answer.
2. **Binary Search on Answer**:
   - Candidate uniqueness values range from 1 to n.
   - For a candidate mid, count how many subarrays have number of distinct elements <= mid.
   - If that count is at least the median position, then the median <= mid; otherwise it is larger.
3. **Counting Subarrays with <= k distinct elements**:
   - Use a sliding window to count subarrays with at most k distinct elements.
   - Maintain a frequency map for the current window.
   - If distinct elements exceed k, move the left pointer until distinct <= k again.
   - Add the number of valid subarrays ending at each right index.
4. **Median Derivation**:
   - total_subarrays = n*(n+1)/2
   - target_rank = (total_subarrays + 1) / 2 (1-indexed)
   - Find the smallest x such that count(distinct <= x) >= target_rank.

### Code
<Implementation reasoning and code following strategy 1...>

<Strategy 2>
### Strategy
1. **Understanding the Problem**:
   - Compute the multiset of distinct-element counts over all subarrays of nums.
   - Return the median of that multiset after sorting.
2. **Key Observations**:
   - There are n*(n+1)/2 subarrays, so enumerating all is infeasible.
   - Instead, count how many subarrays have distinct count <= k for a given k, and use that for selection.
3. **Solution Strategy**:
   - Use a sliding window to compute count_at_most(k).
   - Binary search the smallest k such that count_at_most(k) reaches the median rank.
   - Return that k.

### Code
<Implementation reasoning and code following strategy 2...>
\end{lstlisting}
\end{rawpanel}
\caption{
   Two strategies generated by \method for the same task for the first step in the Coding environment. The text in ``<...>" is for visualization and not part of the actual response. Where specified, some of the response is omitted for clarity. The strategies are under \texttt{<Strategy 1>}, \texttt{<Strategy 2>} and are independently generated and are only sequentially shown for visualization. The two strategies follow different approaches to solving the same task, with strategy 1 directly computing a frequency histogram via enumeration and strategy 2 using binary search and a sliding window. Each strategy is followed by by additional textual reasoning and Python code actually implementing the strategy.
}
\label{fig:strategy_example1}
\end{figure}

\newpage

\begin{figure}[H]
\centering
\begin{rawpanel}
\begin{lstlisting}[style=rawtext]
### Question
You are given a 0-indexed array heights of positive integers, where heights[i] represents the height of the ith building. If a person is in building i, they can move to any other building j if and only if i < j and heights[i] < heights[j]. You are also given another array queries where queries[i] = [ai, bi]. On the ith query, Alice is in building ai while Bob is in building bi. Return an array ans where ans[i] is the index of the leftmost building where Alice and Bob can meet on the ith query. If Alice and Bob cannot move to a common building on query i, set ans[i] to -1.

<Remainder of problem setup...>

Here is my previous FAILED strategy:
```
<Start of strategy omitted...>
Optimization strategy:
- For any query [ai, bi], we need to determine if there's a building j where: - j > max(ai, bi) (since both need to be able to reach j, j must be after both) - height[j] > max(height[ai], height[bi])

We can precompute the next greater element using a stack (monotonic stack) concept to quickly find the next element that is greater than a given value, but since we need the leftmost such building that is greater than both, we can instead iterate from max(ai, bi) + 1 to the end and track the first building with height > max(height[ai], height[bi]).
```

First, after "### Strategy" critique the failed strategy and how it can be fixed. Be precise. Then address this critique by writing a better strategy. Make sure the strategy is detailed and the code is easy to implement from the strategy. Then write the code to solve the question based on the strategy and question in "### Code".

### Strategy
The previous failed approach incorrectly assumes that for a query "[ai, bi]", the meeting building must be to the right of both ai and bi and have height greater than max(heights[ai], heights[bi]). While this condition is necessary, it is not sufficient due to a misunderstanding of the movement constraints.

Critical Insight: The movement rule says that a person from building i can only move to building j if i < j and heights[i] < heights[j]. So Bob is at building 4, and there's no building to his right.

<Remainder of strategy omitted>

### Code
<Implementation reasoning and code following corrected strategy>

\end{lstlisting}
\end{rawpanel}
\caption[Qualitative strategy example]{
  Qualitative example in the Coding environment of how \method uses negative strategy reflection to improve the diversity of generated strategies.
   The text in ``<...>" is for visualization and not part of the actual response. Where specified, some of the response is omitted for clarity.
  The agent reflects on a previously executed strategy that resulted in a failed outcome from earlier in RL training with the negative strategy reflection prompt.
  The original strategy incorrectly missed a corner case by assuming the meeting point must be strictly after both positions. 
  But this misses the case where the rightmost person's building is itself a valid meeting point.
  The agent reflects on this strategy, sees the error, and then addresses the shortcomings of the old strategy with a new strategy.
}
\label{fig:strategy_example2}
\end{figure}
\newpage

\section{Environment Details}
\label{sec:env-details}

\subsection{Coding}
We turn the dataset of coding problems from \citet{xia2025leetcodedataset} into a multi-turn coding environment.
This dataset consists of Python coding problems categorized into ``Easy'', ``Medium'', and ``Hard'' difficulties.
For all of our training, we only work with the 606 problems from the ``Hard'' category.
We report test performance on the 228 hard category tasks from the test set with 8 independent evaluations per episode.

Each problem consists of a problem description, example input and outputs, constraints on input values, and the starter code.
The action space is a Python program, so the agent outputs self-contained Python code to solve the coding problem.
This code is run in a new Python instance with the test conditions provided by the problem episode and must finish within 2 seconds before timing out.
The reward is $+1$ if all of the tests pass within the time limit and $0$ otherwise.
The observation for the next step is the stack trace of any runtime errors that occurred or any of the unit tests that failed.
In a multi-turn fashion, the agent generates a new program based on this observation.
For our experiments, all episodes last for 2 steps, meaning the agent has one action to correct errors from the first action.

\subsection{Android World}
We use the same action and observation space from the Qwen2.5-VL UI agent results reported in \citet{bai2025qwen25vltechnicalreport}. 
The observation space is a $2400 \times 1080$ image.
The action space consists of low-level UI control interactions like tapping an $x,y$ coordinate, scrolling, inputting text, a long tap and more.
We also use the same Qwen2.5-VL UI agent prompt from \citet{bai2025qwen25vltechnicalreport}.
We insert the strategy prompts immediately after the instruction.

We first remove the visual question answering tasks from the original 115 AndroidWorld tasks \cite{rawles2024androidworld}, since we focus on multi-step interactive tasks to achieve a goal rather than training agents to answer questions.
We then removed some of the tasks where the Qwen2.5-VL-72B model achieved no success rate.
We split the remaining 56 tasks into a train and test set for evaluating \method and baselines. 
We use the following 26 tasks for training spanning 11 apps:
\begin{itemize}[itemsep=0pt,topsep=0pt,parsep=0pt,partopsep=0pt,parsep=0pt,leftmargin=*]
  \item AudioRecorderRecordAudioWithFileName
  \item SimpleCalendarAddOneEvent
  \item SimpleCalendarAddOneEventTomorrow
  \item SimpleCalendarAddRepeatingEvent
  \item SimpleCalendarDeleteOneEvent
  \item ContactsAddContact
  \item FilesDeleteFile
  \item MarkorAddNoteHeader
  \item MarkorCreateFolder
  \item MarkorCreateNote
  \item MarkorDeleteNote
  \item MarkorEditNote
  \item SimpleSmsReply
  \item SimpleSmsReplyMostRecent
  \item SimpleSmsSend
  \item RecipeAddMultipleRecipes
  \item RecipeAddSingleRecipe
  \item RecipeDeleteDuplicateRecipes
  \item RecipeDeleteMultipleRecipes
  \item RecipeDeleteSingleRecipe
  \item ExpenseDeleteMultiple
  \item ExpenseDeleteSingle
  \item ExpenseAddSingle
  \item RetroCreatePlaylist
  \item TurnOffWifiAndTurnOnBluetooth
  \item SimpleCalendarAddOneEventRelativeDay
\end{itemize}
We then report the test performance as the average success rate across the following tasks with 4 evaluations per episode.
\begin{itemize}[itemsep=0pt,topsep=0pt,parsep=0pt,partopsep=0pt,parsep=0pt,leftmargin=*]
  \item AudioRecorderRecordAudio
  \item SystemBluetoothTurnOff
  \item SystemBluetoothTurnOn
  \item SystemWifiTurnOff
  \item SystemWifiTurnOn
  \item CameraTakePhoto
  \item CameraTakeVideo
  \item ClockStopWatchRunning
  \item ClockTimerEntry
  \item ExpenseAddMultiple
  \item ExpenseDeleteDuplicates2
  \item ExpenseDeleteMultiple2
  \item FilesMoveFile
  \item MarkorChangeNoteContent
  \item MarkorDeleteAllNotes
  \item MarkorDeleteNewestNote
  \item MarkorMoveNote
  \item RecipeDeleteDuplicateRecipes2
  \item RecipeDeleteDuplicateRecipes3
  \item RecipeDeleteMultipleRecipesWithNoise
  \item RecipeDeleteSingleWithRecipeWithNoise
  \item RetroSavePlaylist
  \item SimpleCalendarAddOneEventInTwoWeeks
  \item SimpleCalendarDeleteEvents
  \item SimpleCalendarDeleteEventsOnRelativeDay
  \item SimpleCalendarEventsInTimeRange
  \item SimpleCalendarNextEvent
  \item SimpleCalendarNextMeetingWithPerson
  \item SystemBrightnessMax
  \item SystemBrightnessMin
\end{itemize}
Notably, the test set includes the following unseen apps: System, Retro, Clock, and Camera. The test set thus includes new tasks in seen apps and new tasks in unseen apps.

To achieve a scalable deployment of AndroidWorld simulations for training and evaluation, we separate the environment from the training or evaluation loops in a dedicated Kubernetes cluster. 
Each Kubernetes pod is equipped with a multi-threaded HTTP server that defines APIs matching those of the Gym environment interface. A light client-environment wrapper is used by the training or evaluation loop to communicate with the environment server to send actions and retrieve environment state (e.g. screenshots). The network latency is minimal compared to the CPU time taken to render the states and run the Android emulator for every step.
Future work can investigate how \method scales to synthetically generated tasks or hybrid action spaces that include tool-calling in AndroidWorld \citep{ramrakhya2025scaling,yang2025ultracua}.

\subsection{Language Rearrangement (LangR)}
Language Rearrangement \cite{szot2023large} is an embodied AI environment using the Habitat simulator \cite{szot2021habitat,puig2023habitat}.
We evaluate the test performance on the ``New Scenes'' split, consisting of 100 testing episodes.
We report numbers with 8 independent evaluations per episode.
The textual observations take the form ``Robot is holding a \texttt{<hold-object>} and is at the \texttt{<match-receptacle>} with objects \texttt{<object1>}, \texttt{<object2>}, \ldots''. 
Where ``\texttt{<hold-object>}'' is the name of the object the robot is holding (such as an ``apple''). 
A receptacle is a static entity that objects can be placed on or within (such as ``fridge'' or ``kitchen table''). 
The ``\texttt{<match-receptacle>}'' is the receptacle that the robot is within 1.5 meters of and is facing.
Finally, ``\texttt{<object1>}'' denotes the names of the objects on the receptacle.

We use the same prompt followed by prior works \citet{szot2023large,szot2024grounding,szot2025multimodal} and add the following text to the end of that prompt to support the textual observations \texttt{Action history:\{history\}\textbackslash n The user query: \{instruction\}\textbackslash n Current robot state: \{state\}\textbackslash n}

\subsection{AppWorld}
AppWorld \citep{trivedi2024appworld} is a simulated app ecosystem with multiple applications and populated synthetic user databases. The agent must output API calls in the form of Python code blocks to fulfill a user instruction, such as using a payments app to send money to friends. 
The agent is allowed to take multiple steps in the environment, observing the output of the previous code block and executing new code blocks.
All code blocks are part of one Python session, meaning that previous variables and definitions can be reused.
The action space is thus a Python code block.
The observation space is the stdout of the previous code block, which includes any print statements or errors.
For our setting, we limit each episode to 16 steps, which we found was sufficient for the Easy split we consider.
We use the same prompt from \citet{trivedi2024appworld}, which includes one in-context example and then describes the rules of the AppWorld APIs.

We train on 36 tasks, which are the ``Easy" category of tasks from the original training set. We then test on the 57 ``Easy" category tasks from the original ``Test Normal" dataset with 8 evaluations per episode.

\applefootnote{ \textcolor{textgray}{\sffamily Apple and the Apple logo are trademarks of Apple Inc., registered in the U.S. and other countries and regions.}}

\end{document}